\documentclass[preprint,3p,authoryear]{elsarticle}

\usepackage{amssymb}
\usepackage{amsmath}
\usepackage{hyperref}
\usepackage{comment}
\usepackage{enumitem}
\usepackage[ruled,vlined]{algorithm2e}
\usepackage{setspace}
\usepackage{float}
\usepackage{booktabs}
\usepackage{url}
\usepackage{caption}
\usepackage{subcaption}
\usepackage{multirow}
\usepackage{tabularx}  
\usepackage{makecell}
\usepackage[labelfont=bf]{caption}

\makeatletter
\def\ps@pprintTitle{%
 \let\@oddhead\@empty
 \let\@evenhead\@empty
 \let\@oddfoot\@empty
 \let\@evenfoot\@empty
}
\makeatother

\date{}

\begin{document}

\begin{frontmatter}

\vspace*{1cm} 

\begin{center}
    {\fontsize{15pt}{18pt}\selectfont
RocketStack: Level-aware Deep Recursive Ensemble Learning Architecture
}\\[15pt]

    {\normalsize \c{C}a\u{g}atay Demirel$^{\text{a,*}}$}\\[3pt]

    \vspace{1em}
    $^a$\textit{\small Donders Institute for Brain, Cognition and Behaviour, Kapittelweg 29, Nijmegen, 6525 EN, Netherlands}\\
    \vskip 2mm
    $^*$\small Correspondance: cagatay.demirel.sci@gmail.com
\end{center}

\begin{abstract}

\noindent
Ensemble learning remains a cornerstone of machine learning, with stacking used to integrate predictions from multiple base learners through a meta-model. However, deep stacking remains uncommon due to feature redundancy, complexity, and computational burden. To address these limitations, RocketStack is introduced as a level-aware recursive stacking architecture explored up to ten stacking levels, extending beyond prior architectures. At level 1, base-learner predictions are fused with original features; at later levels, weaker learners are incrementally pruned using out-of-fold (OOF) scores. To curb early saturation, pruning is regularized by applying Gaussian perturbations at two noise scales to OOF scores prior to model selection for next-level stacking, alongside deterministic pruning. To control feature growth, periodic compression is applied at levels 3, 6, and 9 using  Simple, Fast, Efficient (SFE) filtering, attention-based selection, and autoencoders. Across 33 datasets (23 binary, 10 multi-class), increasing accuracy with depth is confirmed by linear mixed-effects trend tests, and the best meta-model per level increasingly outperforms the best standalone ensemble. OOF-perturbed pruning is found to improve stability and late-level gains, while periodic compression is found to yield substantial runtime and dimensionality reductions with minimal accuracy drop. At the deepest level, accuracy slightly surpasses established deep tabular baselines. When hyperparameter optimization is performed on baseline models, early performance is boosted; however, untuned RocketStack closes the gap with depth and remains competitive at later levels. It achieves deep recursive stacking with sublinear computational growth and provides a modular, depth-aware foundation for scalable decision fusion as model pools and feature spaces evolve.

\end{abstract}


\begin{keyword}
stack ensemble \sep
blend ensemble \sep
ensemble learning \sep
deep stacking \sep
stochastic perturbation \sep
feature fusion \sep
attention layer \sep
meta-classifier \sep
hyperparameter optimization
\end{keyword}

\end{frontmatter}

\begin{figure}[H]
 \centering \includegraphics[width=1\textwidth,keepaspectratio=false]{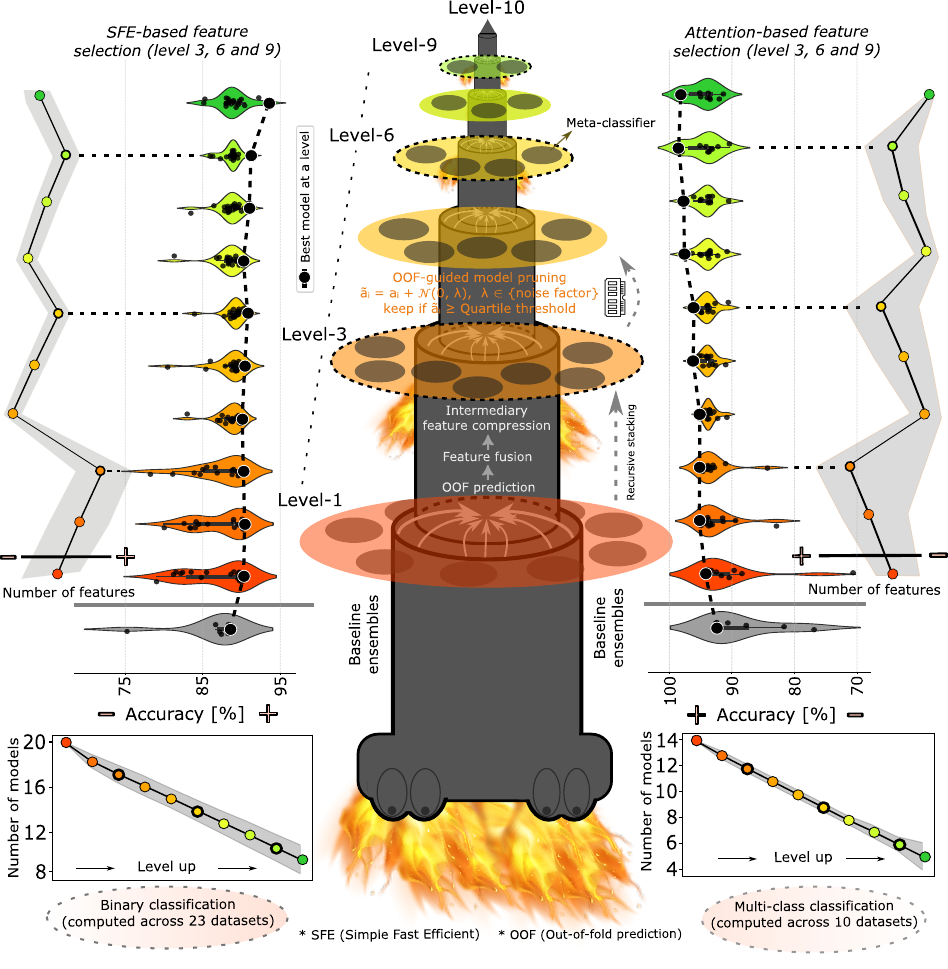}
 \caption*{Graphical abstract illustrates RocketStack framework} 
  \label{fig:graphicalabstract} 
\end{figure}

\section{Introduction}




Ensemble learning has become a cornerstone of modern machine learning, known for its capacity to reduce variance, mitigate overfitting, and improve predictive robustness across diverse domains especially in structured tabular data where deep neural networks often underperform \cite{b1_zhou2012ensemble}. Among ensemble strategies, stacking (also known as stacked generalization) stands out as a meta-learning paradigm that combines heterogeneous base models via a higher-order learner \cite{b2_wolpert1992stacked, b3_breiman1996stacked}. This architecture enables the capture of complementary decision boundaries, and has demonstrated superior performance in benchmarks, industrial applications, and competitive platforms like Kaggle \cite{b4_bojer2021kaggle}. Its success is particularly pronounced in high-dimensional tabular settings, where handcrafted feature sets dominate, and where stacking often forms the foundation of top-performing pipelines.


Despite its empirical appeal, stacking remains predominantly shallow in practice typically restricted to one or two meta-layers. The key barriers to deeper adoption include feature accumulation, training inefficiency, and model overfitting. As predictions are propagated through stacking layers, dimensionality increases, leading to bloated feature spaces that degrade generalization performance \cite{b5_wang2023multilayer}. Runtime costs also escalate, especially when computationally expensive learners are used without coordinated pruning or compression \cite{b6_adam2024ensemble}. Moreover, recursive application of nonlinear learners compounds overfitting risks, particularly in low-sample settings or when cross-validation strategies are not tightly integrated \cite{b7_rane2024ensemble}. These limitations have curtailed widespread experimentation with deep stacking, despite its theoretical capacity for hierarchical representation learning.



Recent work has addressed stacking complexity by combining SHAP-based interpretability with Recursive Feature Elimination (RFE) to identify salient features within ensemble pipelines, improving both interpretability and performance \cite{b8_huang2023rfe}. Another study have incorporated RFE directly into stacking architectures to filter redundant meta-features and reduce training cost \cite{b9_idris2024stacking}. In parallel, ensemble pruning and selection strategies have emerged to reduce redundancy and complexity before meta-fusion. For instance, \cite{b10_aljamaan2024dynamic} explores greedy and backward-elimination methods to pre-select base models for code smell detection, while \cite{b11_demirel2023200185} presents two strategies: out-of-fold (OOF)-based weighting for probabilistic averaging of model outputs, and model pruning followed by meta-level stacking to boost ad click prediction performance. However, most existing approaches operate only at shallow depths and lack mechanisms to coordinate information across successive levels, rendering truly deep stacking impractical.

\subsection{Motivation and contribution}



While ensemble learning methods have achieved widespread success, the development of deep recursive stacking architectures remains nascent, largely due to persistent challenges such as feature accumulation, runtime complexity, and model overfitting. These issues are especially pronounced in high-dimensional tabular domains, where uncontrolled propagation of intermediate features across ensemble levels can lead to redundancy, bloated training times, and unstable generalization \cite{b12_ganaie2022ensemble}. Although recent studies have explored strategies like dimensionality reduction or level-specific pruning, these efforts typically operate in isolated stages and lack the modular scalability required to maintain efficiency across recursive ensemble depths \cite{b14_du2025foundations}. This architectural bottleneck underscores a need for ensemble models that incorporate depth-aware optimization, runtime pruning, and adaptive feature control to enable truly scalable stacking.

\begin{figure*}
 \centering \includegraphics[width=0.8\textwidth,keepaspectratio=false]{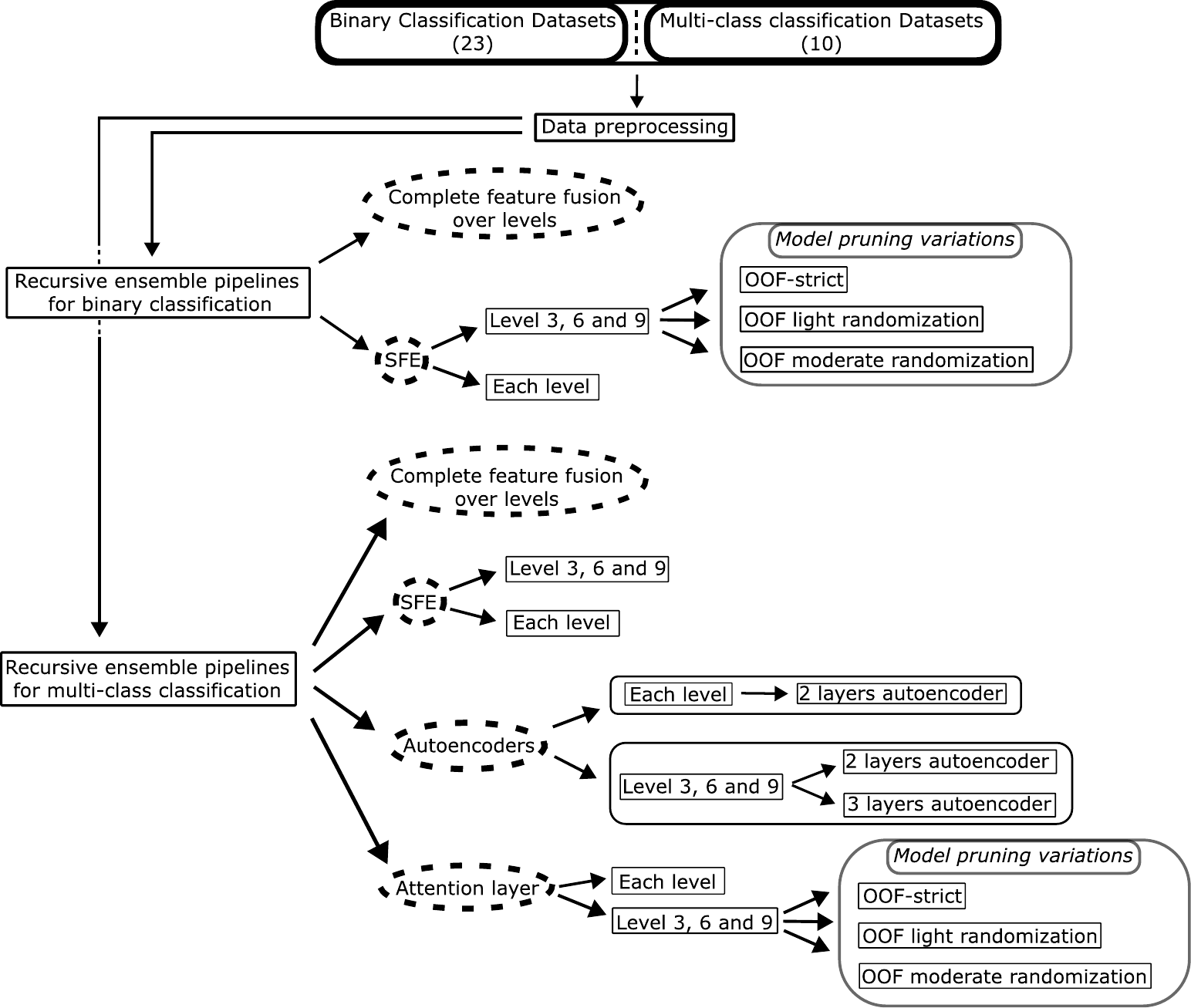}
 \caption{Schematic overview of the study} 
  \label{fig:overviewofthesystem} 
\end{figure*}

To address the architectural and computational limitations of existing stacking models, this work introduces RocketStack, a modular system for recursive ensemble learning with scalable depth and structured optimization. The schematic overview of the whole study is illustrated in \autoref{fig:overviewofthesystem}. This work makes the following key contributions:

\begin{enumerate}[label=(\roman*)]

    \item A modular recursive-stacking architecture, RocketStack, is introduced and explored to depth 10, with built-in controls that limit model redundancy, feature growth, and runtime.

    \item Dynamic pruning is implemented by applying percentile OOF-score thresholds, with mild Gaussian noise injected into the scores to prevent premature convergence and enhance accuracy.

    \item Alongside a no-compression baseline, three feature compression schemes—SFE \cite{b14_1_sfe}, autoencoders \cite{b14_2_autoencoders}, and attention mechanisms \cite{b14_3_attention}—are evaluated under both per-level and periodically (at levels 3, 6, 9); periodic compression is shown to curb feature inflation while improving the trade-off between accuracy and runtime.

    \item Consistent performance gains across stacking depths are observed in 33 binary and multi-class datasets and across RocketStack variants, as confirmed by trend tests and level-wise comparisons.

    \item The advantage of base-level hyperparameter optimization (HPO) is shown to diminish with stacking depth, as untuned RocketStack configurations at deeper levels progressively narrow the gap to their HPO tuned counterparts.

\end{enumerate}

\section{Related work}
\subsection{Classical ensemble learning in tabular contexts}


Ensemble learning methods such as bagging, boosting, and stacking have long served as foundational strategies in predictive modeling, particularly within structured tabular domains where data lacks spatial or sequential inductive biases. Bagging reduces variance by aggregating bootstrapped base learners in parallel \cite{b15_breiman1996bagging}, while boosting models like AdaBoost \cite{b16_freund1997decision} and gradient boosting \cite{b17_friedman2001greedy} iteratively refine weak learners to reduce bias through sequential corrections. Stacking further expands this paradigm by leveraging meta-learning to integrate predictions from heterogeneous base models, thereby capturing diverse predictive patterns \cite{b2_wolpert1992stacked, b3_breiman1996stacked}. These ensemble strategies remain central to contemporary architectures particularly in tabular machine learning pipelines where they often outperform deep learning models due to their inductive alignment with structured feature spaces.

In these domains, where data structures are typically non-hierarchical and feature relationships are not spatially encoded, ensemble methods especially tree-based boosting algorithms have consistently outperformed deep neural networks. This advantage stems from their ability to handle mixed data types, require less data, and capture non-linear interactions effectively \cite{b20_gorishniy2021revisiting}. Benchmark studies show that models like XGBoost, LightGBM, and CatBoost dominate structured tasks in AutoML frameworks and competitive platforms such as Kaggle \cite{b22_tsch2024datacentric}. 

However, classical ensemble designs in these settings tend to favor shallow, flat architectures that optimize learner diversity at a single level but lack recursive depth. Even in AutoML pipelines, stacked ensembles are often implemented with fixed depth and heuristic-level selection, limiting their scalability \cite{b25_chen2021automl}. Without structured pruning or inter-level coordination, such models risk computational inefficiency and representational redundancy when extended across levels \cite{b26_alsaffar2024enhancing}. Thus, despite their empirical strength, ensemble models in tabular contexts remain constrained by architectural shallowness and the absence of depth-aware optimization models.

\subsection{Recursive and multi-level stacking}

Stacking, originally conceptualized by \cite{b2_wolpert1992stacked} and later formalized for regression by \cite{b3_breiman1996stacked}, remains a foundational ensemble technique. Despite its appeal, most AutoML implementations retain shallow or template-based stacking, largely constrained by time and resource budgets \cite{b27_ferreira2021comparison}. Early attempts to move beyond classical stacking have reported only marginal gains, likely due to limited recursive control or coordinated optimization. For example, \cite{b11_demirel2023200185} compared level-1 stacking with a blend architecture in which meta-learner predictions were concatenated with the original features, but no significant improvement was observed over standard stacking, suggesting that single meta-level is insufficient and motivating deeper, recursively structured stacking.

Efforts to scale stacking into hierarchical ensembles repeatedly encounter runtime cost, representational redundancy, and overfitting, particularly when depth is increased without structural coordination or pruning \cite{b12_ganaie2022ensemble, b7_rane2024ensemble}. Surveys similarly note that most pipelines remain shallow or fixed-depth due to limited adaptivity and elevated overfitting risk \cite{b28_mienye2022survey}, and even optimized application-specific stacking (e.g., medical diagnosis or cybersecurity) shows limited modular scalability and little benefit beyond two meta-levels \cite{b29_kumar2022optimized}. These findings point to a broader architectural gap: a principled recursive ensemble framework that is both modular and computationally scalable remains lacking. Although meta-learning increasingly emphasizes recursive dynamics, level-adaptive deep stacking is still uncommon in practice; for instance, \cite{b30_zhao2022autodes} introduced AutoDES for dynamic ensemble selection in AutoML, but its control logic remains sequential rather than recursively optimized.

\subsection{Feature selection within ensemble learning}

Feature selection plays a critical role in ensemble learning by improving interpretability, reducing overfitting, and increasing computational efficiency, particularly in multilayered architectures such as stacking. Selection techniques are commonly grouped into filter methods (e.g., mutual information, correlation coefficients), wrapper methods (e.g., Boruta, RFE), and embedded strategies (e.g., LASSO, tree importance). In stacking-based ensembles, hybrid combinations of these approaches have been reported to yield substantial performance benefits in domains such as medical diagnosis and cybersecurity \cite{b33_alsaffar2024shielding}. Wrapper-based methods including RFE and Boruta have been extensively paired with ensemble models to stabilize feature subsets across cross-validation folds and hierarchical tree structures \cite{b36_habibi2023hybrid}. In parallel, attention mechanisms have been adopted in deep learning for dynamic modeling of feature relevance \cite{b40_kumar2024transformer}, although their integration into recursive or layered ensemble learning remains comparatively underexplored.

Advanced ensemble pipelines increasingly incorporate hybrid selectors such as SHAP-RFE \cite{b39_zhu2025shaprf} and rank-based ensemble selectors to improve stability and interpretability, yet these methods are typically applied as preprocessing or deployed within isolated ensemble layers, rather than being coordinated recursively across depth. Such localized selection can amplify redundancy across stacked layers and inflate computational overhead through unregulated dimensional overlap. Despite promising task-specific gains, generalized frameworks that couple feature selection with runtime-aware feedback across recursive ensemble depths remain limited. RocketStack addresses this gap by applying exploratory feature selection either at every level or at periodic depths and pairing it with adaptive pruning to keep the meta-feature space compact and computationally efficient throughout the hierarchy.

\subsection{Pruning and complexity control in ensembles}




Pruning is a critical mechanism for curbing model proliferation, lowering runtime, and limiting overfitting effects that are amplified in multi-level stacks: static schemes rely on preset accuracy or importance thresholds, whereas dynamic schemes use validation feedback to decide which learners to retain \cite{b42_shen2023dynamic}. Recent work has further framed pruning as an explicit optimization problem \cite{b44_wu2024hierarchical}, for example by leveraging focal diversity to adaptively trim hierarchical structures, collectively underscoring that pruning is most effective when treated as an integral design element rather than a post-hoc fix. Cost-sensitive variants extend this perspective by coupling selection decisions to runtime or latency budgets, such as multifidelity stacking, where learners are added only when runtime-guided error bounds justify the cost \cite{b45_sung2022stackingdesigns}, or greedy rules that prune early weak learners to reduce inference delay \cite{b46_grubb2012speedboost}. In line with these principles, RocketStack retains models only when their out-of-fold (OOF) scores exceed an adaptive percentile threshold, and applies this recursive pruning at fixed depths trimming both computation and feature growth as stacking depth increases.

\section{Methodology}

This section outlines the design and evaluation of the proposed RocketStack architecture, developed for structured tabular datasets in both binary and multi-class classification contexts. The methodology emphasizes a recursive stacking architecture that integrates feature selection and pruning strategies across varying ensemble depths. In particular, three distinct fusion strategies are explored: (i) level-wise feature selection applied at every layer, (ii) periodic feature selection at fixed depths (levels 3, 6, and 9), and (iii) a baseline \textit{vanilla fusion} approach with no intermediate selection. These configurations are evaluated to assess the impact of feature curation frequency on model performance and structural efficiency. The following subsections provide detailed descriptions of the ensemble construction process, the set of baseline classifiers employed, and the fusion and pruning mechanisms implemented throughout the RocketStack pipeline.

\subsection{Dataset selection and preprocessing}

To evaluate the performance and generalizability of the proposed RocketStack architecture, a total of 33 publicly available datasets were selected from OpenML repository. This includes 23 binary classification datasets (credit-g, banknote-authentication, Satellite, churn, steel-plates-fault, autoUniv-au1-1000, qsar-biodeg, pc3, phoneme, ozone-level-8hr, hill-valley, spambase, kc1, pc1, pc4, wilt, blood-transfusion-service-center, wdbc, kc2, climate-model-simulation-crashes, breast-w, diabetes, credit-approval) and 10 multi-class classification datasets (mfeat-morphological, segment, cmc, optdigits, page-blocks, mfeat-factors, balance-scale, cardiotocography, JapaneseVowels, pendigits). These datasets span a wide range of domains, including finance, healthcare, industrial engineering, software defect prediction, environmental modeling, biology and chemical assay classification, speech and handwriting recognition, document analysis, and demographic studies. The inclusion of datasets from such diverse sources ensures a rigorous and domain-independent evaluation of the recursive ensemble architecture. 

All datasets went through preprocessing to ensure compatibility with ensemble learners requiring fully numerical input. For each dataset, categorical features identified automatically or optionally provided were encoded using one-hot encoding with the OneHotEncoder transformer, while numerical features were retained as-is. A ColumnTransformer was used to apply these transformations efficiently and consistently across datasets. Target labels were encoded into integer values using LabelEncoder to standardize class representation across binary and multi-class datasets. The final feature matrices were converted to dense format for compatibility with model training pipelines. This preprocessing pipeline ensured that all datasets, regardless of their original format or feature types, were transformed into a uniform, fully numerical structure.

A complete summary of the selected datasets including the number of samples, features, classes, and source references is provided in Table~\ref{tab:datasets}. For clarity, the datasets are divided into two categories: binary classification datasets are listed in Table~\ref{tab:datasets-a}, and multi-class classification datasets are shown in Table~\ref{tab:datasets-b}.

\begin{table}
\caption{Overview of the datasets used in this study}
\label{tab:datasets}
\centering
\subfloat[Binary classification 23 datasets\label{tab:datasets-a}]{
\fontsize{6}{10}\selectfont
\begin{tabular}{|l|r|r|r|l|}
\hline
\textbf{Dataset name} & \textbf{\#Samples} & \textbf{\#Features} & \textbf{\#Classes} & \textbf{Reference} \\
\hline
credit-g & 1000 & 61 & 2 & \url{https://www.openml.org/data/download/31/dataset_31_credit-g.arff} \\
banknote-authentication & 1372 & 4 & 2 & \url{https://www.openml.org/data/download/1586223/php50jXam} \\
Satellite & 5100 & 36 & 2 & \url{https://www.openml.org/data/download/16787463/phpZrCzJR} \\
churn & 5000 & 33 & 2 & \url{https://www.openml.org/data/download/4965302/churn.arff} \\
steel-plates-fault & 1941 & 33 & 2 & \url{https://www.openml.org/data/download/1592296/php9xWOpn} \\
autoUniv-au1-1000 & 1000 & 20 & 2 & \url{https://www.openml.org/data/download/1593743/php7KLval} \\
qsar-biodeg & 1055 & 41 & 2 & \url{https://www.openml.org/data/download/1592286/phpGUrE90} \\
pc3 & 1563 & 37 & 2 & \url{https://www.openml.org/data/download/53933/pc3.arff} \\
phoneme & 5404 & 5 & 2 & \url{https://www.openml.org/data/download/1592281/php8Mz7BG} \\
ozone-level-8hr & 2534 & 72 & 2 & \url{https://www.openml.org/data/download/1592279/phpdReP6S} \\
hill-valley & 1212 & 100 & 2 & \url{https://www.openml.org/data/download/1593762/phpoW7Dbi} \\
spambase & 4601 & 57 & 2 & \url{https://old.openml.org/data/download/22103191/dataset} \\
kc1 & 2109 & 21 & 2 & \url{https://www.openml.org/data/download/53950/kc1.arff} \\
pc1 & 1109 & 21 & 2 & \url{https://www.openml.org/data/download/53951/pc1.arff} \\
pc4 & 1458 & 37 & 2 & \url{https://www.openml.org/data/download/53932/pc4.arff} \\
wilt & 4839 & 5 & 2 & \url{https://www.openml.org/data/download/18151926/phpcSeK3V} \\
blood-transfusion & 748 & 4 & 2 & \url{https://api.openml.org/data/download/22125224/dataset} \\
wdbc & 569 & 30 & 2 & \url{https://www.openml.org/data/download/1592318/phpAmSP4g} \\
kc2 & 522 & 21 & 2 & \url{https://www.openml.org/data/download/53946/kc2.arff} \\
climate-model-crashes & 540 & 20 & 2 & \url{https://www.openml.org/data/download/1586232/phpXeun7q} \\
breast-w & 699 & 9 & 2 & \url{https://www.openml.org/data/download/52350/openml_phpJNxH0q} \\
diabetes & 768 & 8 & 2 & \url{https://www.openml.org/data/download/22044302/diabetes.arff} \\
credit-approval & 690 & 51 & 2 & \url{https://www.openml.org/data/download/29/dataset_29_credit-a.arff} \\
\hline
\end{tabular}
}
\vspace{1em}

\subfloat[Multi-class classification 10 datasets\label{tab:datasets-b}]{
\fontsize{6}{10}\selectfont
\begin{tabular}{|l|r|r|r|l|}
\hline
\textbf{Dataset name} & \textbf{\#Samples} & \textbf{\#Features} & \textbf{\#Classes} & \textbf{Reference} \\
\hline
mfeat-morphological & 2000 & 6 & 10 & \url{https://www.openml.org/data/download/18/dataset_18_mfeat-morphological.arff} \\
segment & 2310 & 18 & 7 & \url{https://www.openml.org/data/download/22045435/segment.arff} \\
cmc & 1473 & 24 & 3 & \url{https://www.openml.org/data/download/23/dataset_23_cmc.arff} \\
optdigits & 5620 & 64 & 10 & \url{https://www.openml.org/data/download/53514/optdigits.arff} \\
page-blocks & 5473 & 10 & 5 & \url{https://www.openml.org/data/download/53555/page-blocks.arff} \\
mfeat-factors & 2000 & 216 & 10 & \url{https://www.openml.org/data/download/12/dataset_12_mfeat-factors.arff} \\
balance-scale & 625 & 4 & 3 & \url{https://www.openml.org/data/download/53531/balance-scale.arff} \\
cardiotocography & 2126 & 35 & 10 & \url{https://www.openml.org/data/download/1593756/phpW0AXSQ} \\
JapaneseVowels & 9961 & 14 & 9 & \url{https://www.openml.org/data/download/53510/kdd_JapaneseVowels.arff} \\
pendigits & 10992 & 16 & 10 & \url{https://www.openml.org/data/download/53553/pendigits.arff} \\
\hline
\end{tabular}
}
\end{table}

\subsection{Baseline classifiers}

To ensure robust comparative evaluation, a wide range of well-established classifiers were selected as level-0 learners in both binary and multi-class settings. For binary classification, 20 classifiers were utilized: XGBoost, LightGBM, Random Forest, Support Vector Classifier (SVC), Stochastic gradient descent (SGD), Catboost, Bagging, AdaBoost, k-Nearest Neighbors (KNN), Linear Discriminant Analysis (LDA), Extra Trees, Gradient Boosting, Calibrated Ridge, Logistic Regression, Calibrated Passive Aggressive, Bernoulli Naive Bayes (BernoulliNB), Gaussian Naive Bayes (GaussianNB), Multi-Layer Perceptron (MLP), and HistGradientBoosting. For multi-class classification tasks, a reduced set of 14 classifiers was employed. SGD, LDA and BernoulliNB were excluded due to low performance at level-0, which raised concerns about its capacity to contribute meaningfully to deeper ensemble layers. Additionally, the calibrated variants of Ridge and Passive Aggressive classifiers implemented via CalibratedClassifierCV to enable probabilistic outputs were excluded due to their computational cost. These models internally perform cross-validation for calibration, which substantially inflates runtime, especially in multi-class settings where output dimensionality increases with the number of classes. CatBoost was also excluded due to its high computational cost in the multi-class setting. The full list of models along with their hyperparameter configurations is provided in Table~\ref{tab:model_hyperparams_framed}. Furthermore, to assess the impact of base learner optimization on the overall RocketStack architecture, additional experiments were conducted by applying Bayesian HPO with 50 iterations per training fold \cite{b11_demirel2023200185, Victoria2021} to the level-0 classifiers in selected binary and multi-class RocketStack settings.

Selected classifiers represent a blend of tree-based ensembles, linear models, kernel-based learners, probabilistic methods, and neural networks each widely validated in prior benchmarking studies for both binary and multi-class classification tasks \cite{choudhury2024scenarios}. In particular, XGBoost, LightGBM, and Random Forest have consistently demonstrated state-of-the-art performance in ensemble pipelines \cite{choudhury2024scenarios}, while MLP, Gaussian NB, and Logistic Regression provide valuable diversity in decision boundaries and learning biases.

\begin{figure*}
 \centering \includegraphics[width=1\textwidth,keepaspectratio=false]{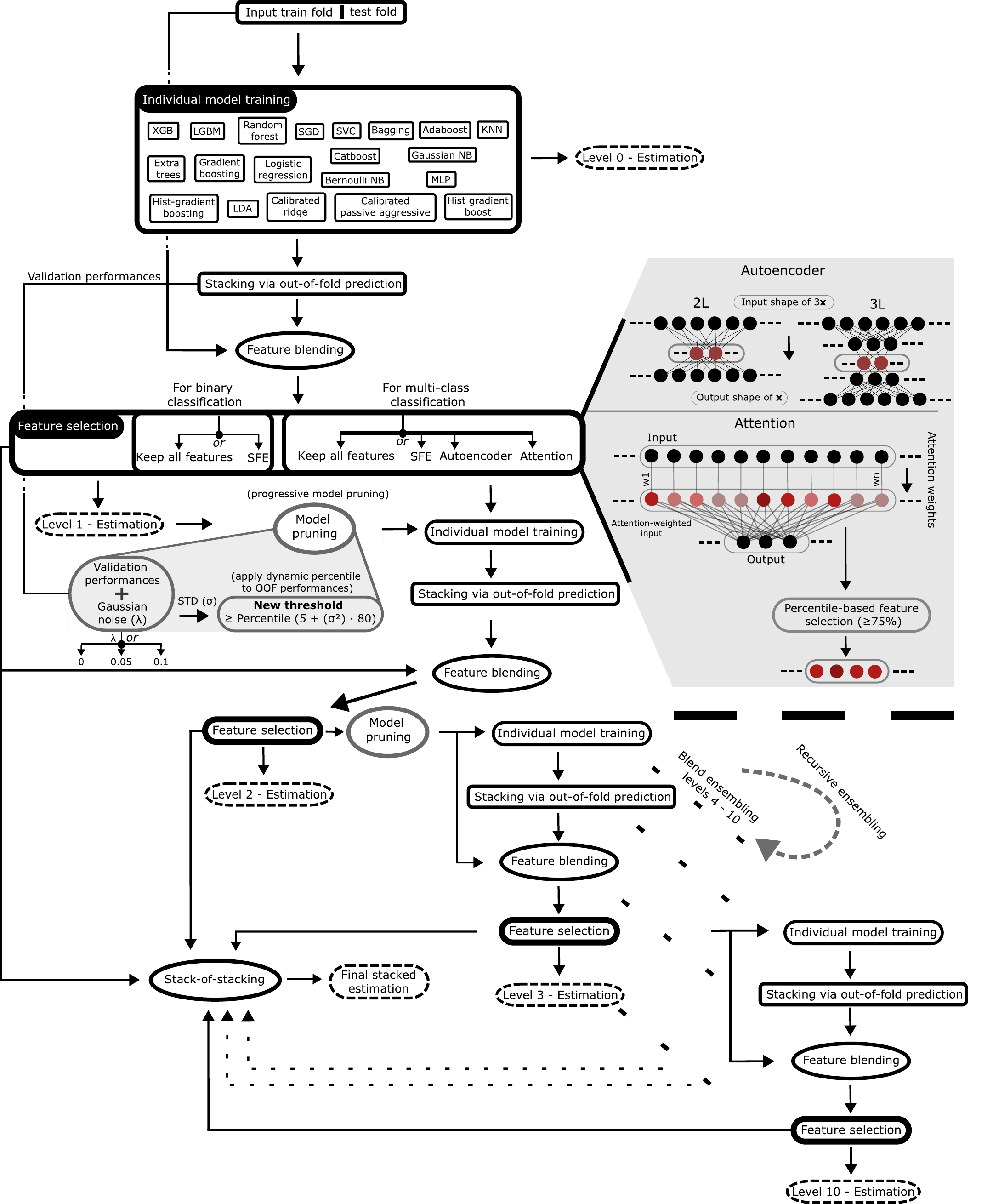}
 \caption{Multi-level recursive ensemble learning (RocketStack) pipeline} 
  \label{fig:ensemblingpipeline_binaryandmc_together} 
\end{figure*}

\subsection{Feature fusion and compression}

In RocketStack, every stacking level starts with OOF probabilities derived from a 5-fold cross-validation loop: each base learner is trained on four folds and predicts on the held-out fold, so every training case receives a score from a model that has never seen it. The OOF vectors from all learners are concatenated to form a meta-feature matrix that feeds the next level. In binary tasks this matrix grows slowly, but in multi-class tasks its width scales with both the number of models and the number of classes, and can balloon at deeper levels. Feature selection or compression is therefore inserted either at every level or periodically (levels 3, 6, 9) to limit dimensionality without discarding useful information. Level 1 blends the OOF probabilities with the original input features; each later level blends new OOF scores with the already compressed matrix from the previous level. Whenever selection is scheduled, blending is followed immediately by dimensionality reduction, and the compressed matrix is passed on, controlling feature growth and runtime while retaining the signal needed for deeper stacking.

For binary classification, three configurations are supported: no compression, in which all features are retained throughout, per-level SFE selection, applied independently at every level, and periodic SFE selection, applied only at levels 3, 6, and 9. The periodic setting allows limited feature accumulation between recalibration steps and constrains overfitting and runtime. For multi-class classification, a broader pipeline is evaluated, comprising eight variants. Besides the no-compression baseline and two SFE options (per-level and periodic), autoencoder-based compression is tested in three forms: a two-layer network run at every level, the same two-layer network run periodically, and a deeper three-layer network run at the same checkpoints. In each case, the current feature matrix is compressed to one-third of its width regardless of its growth. A final family uses an attention layer: after training, features with relevance scores above the 75th percentile are retained and the rest discarded, so the subset size adapts automatically rather than being fixed.

Across both pipelines, feature selection is applied prior to model pruning within each recursive stage to ensure that retained meta-features remain compact and contextually meaningful. A detailed flow chart of the architecture is provided in \autoref{fig:ensemblingpipeline_binaryandmc_together}. All feature selection methods are described below.

\subsubsection{SFE: Simple, Fast, and Efficient Feature Selection}

The SFE method applies a greedy, utility-driven approach to select features based on their relevance to the target and redundancy with already selected features. The core utility function used to score each feature is defined in Eq.~\eqref{eq:sfe}:

\begin{equation}
\mathcal{U}(f) = \frac{\mathrm{Rel}(f)}{1 + \mathrm{Red}(f)},
\label{eq:sfe}
\end{equation}

where $\mathrm{Rel}(f)$ measures the relevance of feature $f$ to the target variable, and $\mathrm{Red}(f)$ quantifies the redundancy with features already included in the selection set. Features are added iteratively to maximize $\mathcal{U}(f)$ until a utility threshold or a predefined feature count is reached. This method is lightweight and highly scalable for recursive pipelines.

\vspace{0.5em}

\subsubsection{Autoencoder-Based Dimensionality Reduction}

Autoencoders offer a non-linear mechanism to compress high-dimensional inputs by learning low-dimensional latent representations. The transformation involves an encoder-decoder pair, where the reconstruction process is described in Eq.~\eqref{eq:ae_recon}:

\begin{equation}
\hat{X} = g_\phi(f_\theta(X)), \quad \text{with} \quad f_\theta: \mathbb{R}^d \rightarrow \mathbb{R}^k, \quad k < d,
\label{eq:ae_recon}
\end{equation}

and the optimization objective is defined in Eq.~\eqref{eq:ae_loss}:

\begin{equation}
\mathcal{L}_{\text{AE}} = \| X - \hat{X} \|^2.
\label{eq:ae_loss}
\end{equation}

In this setup, $f_\theta(X)$ denotes the compressed feature set and $g_\phi$ attempts to reconstruct the original input. The dimensionality $k$ of the latent space is typically set to one-third of the original feature space $d$. Autoencoders are trained to minimize reconstruction loss, promoting efficient encoding of latent structure while discarding noise and redundancy. Two autoencoder configurations are used for feature compression in multi-class RocketStack: a 2-layer autoencoder (2L), consisting of a single encoding layer that reduces the input dimensionality to one-third of its original size, followed by a symmetric decoder; and a 3-layer autoencoder (3L), which uses two successive encoding layers that reduce the input to one-half and then to one-third of its original size, followed by a mirrored decoding path. Both variants are trained using mean squared error loss.

\vspace{0.5em}

\subsubsection{Attention-Based Feature Weighting}

The attention mechanism learns a relevance score for each feature through a trainable transformation of the input. As shown in Eq.~\eqref{eq:att_score}, an attention score vector $\alpha = [\alpha_1, \alpha_2, \ldots, \alpha_d]$ is computed from the input vector $X = [x_1, x_2, \ldots, x_d]$:

\begin{equation}
\alpha = \text{softmax}(W X + b),
\label{eq:att_score}
\end{equation}

where $W$ and $b$ are learnable parameters. The softmax ensures that attention weights are non-negative and sum to one, facilitating interpretability as feature relevance probabilities.

To perform feature selection based on these learned weights, a percentile-based masking operation is applied. Specifically, as formalized in Eq.~\eqref{eq:att_mask}, only features whose attention scores fall within the top 25\textsuperscript{th} percentile of the distribution are retained:

\begin{equation}
X^{\text{att}} = \{x_i \in X \;|\; \alpha_i \geq Q_{75}(\alpha)\},
\label{eq:att_mask}
\end{equation}

where $Q_{75}(\alpha)$ denotes the 75\textsuperscript{th} percentile of the attention score vector. This filtering mechanism enables selective emphasis on the most informative features, as determined by the model’s own learned focus, rather than arbitrary dimensionality reduction.

\subsection{Model pruning via OOF-guided model performances}

In RocketStack, model pruning serves as a core regulatory mechanism for ensemble complexity by controlling the number of active learners and the recursive propagation of predictions. Whereas feature selection regulates feature-space dimensionality, pruning regulates the learner space itself. This control is applied in both binary and multi-class settings, and is particularly important in multi-class configurations where recursive concatenation of OOF predictions can otherwise lead to severe dimensional expansion. Pruning is applied consistently at each recursive level starting from level 1, always as the final operation within the level. When feature selection is present, pruning follows it; otherwise, it is applied directly after feature blending. This enforces a structured reduction of the model space before the next stacking iteration, ensuring that only models with validated performance contribute downstream. Selection is based on OOF validation scores, where only models exceeding a dynamic percentile threshold computed using both the central tendency and variance of model-wise accuracies are retained. 

Three pruning modes are explored. The default mode, OOF-strict, applies deterministic pruning using the raw OOF scores and is used across all feature selection strategies and datasets. Two stochastic variants inject Gaussian noise into OOF scores prior to thresholding, using a light randomization factor ($\lambda = 0.05$) or a moderate factor ($\lambda = 0.1$), thereby relaxing selection boundaries to encourage greater ensemble diversity. These noise-based variants are evaluated only in targeted exploratory settings, namely on the top-performing feature selection strategy in each task domain: periodic SFE at levels 3, 6, and 9 for binary classification, and periodic attention-based selection at levels 3, 6, and 9 for multi-class classification. This design enables analysis of how pruning stochasticity interacts with already promising compression strategies without incurring the computational burden of applying randomization across all conditions. The pruning formulation is provided in Eq.~\eqref{eq:blurred_oof}.

\begin{equation}
\tilde{a}_i = a_i + \epsilon_i, \quad \epsilon_i \sim \mathcal{N}\left(0, \lambda \cdot \mathrm{range}(\mathbf{a})\right)
\label{eq:blurred_oof}
\end{equation}

where $a_i$ denotes the out-of-fold (OOF) score of the $i$-th model, and $\tilde{a}_i$ is its noise-perturbed counterpart used during model selection. The additive noise term $\epsilon_i$ is drawn from a zero-mean Gaussian distribution with standard deviation $\sigma = \lambda \cdot \mathrm{range}(\mathbf{a})$, where $\lambda \in \{0,\, 0.05,\, 0.1\}$ modulates the noise intensity. The term $\mathrm{range}(\mathbf{a}) = \max(\mathbf{a}) - \min(\mathbf{a})$ quantifies the score dispersion across all candidate models and acts as a scaling factor for the noise. The blurred scores $\tilde{\mathbf{a}} = \{\tilde{a}_1, \dots, \tilde{a}_n\}$ are then used to compute the custom pruning threshold $Q_{\mathrm{custom}}$ in Eq.~\eqref{eq:qcustom}.

\begin{equation}
Q_{\mathrm{custom}} = \mathrm{percentile}\left(\tilde{\mathbf{a}},\; 5 + 80 \cdot \mathrm{std}(\tilde{\mathbf{a}})^2 \right)
\label{eq:qcustom}
\end{equation}

where models satisfying $\tilde{a}_i \ge Q_{\mathrm{custom}}$ are retained for the next recursive level.

\vskip 2mm

In all modes, pruning respects fold-wise data boundaries, maintaining clean separation between training and testing during cross-validation. If the number of retained models at a level falls below a predefined lower bound, the recursion halts to preserve structural integrity. Otherwise, the selected models are propagated forward, maintaining a memory-informed recursive composition of the ensemble.

This pruning protocol (whether based on strict OOF performance or exploratory noise-infused variation) serves two roles within RocketStack. First, it regulates ensemble complexity, preventing the uncontrolled accumulation of model outputs across levels. Second, the stochastic variants act as a controlled perturbation of the model selection heuristic, allowing less dominant learners a probabilistic chance to persist. This exploratory mechanism draws inspiration from regularization principles, where selective uncertainty may help downstream estimators avoid overfitting to early-layer dynamics or converging prematurely to suboptimal ensemble compositions. The pseudocode of RocketStack pipeline is formalized in Algorithm~\ref{alg:rocketstack_pseudocode}.

All experiments and model evaluations were conducted on a computer equipped with an Intel Core i7-9700KF (3.60~GHz) processor, 128~GB of DDR4 RAM, and an NVIDIA RTX~2070~Super (8~GB VRAM). To address computational complexity, assuming a typical base-learner training cost $C(N,D)\approx\mathcal{O}(N \cdot D\ \cdot log N)$, a loose worst-case bound without pruning or compression is $\mathcal{O}(L \cdot M \cdot N \cdot D \cdot \log N)$. Under RocketStack, however, the effective training cost is better characterized as $\mathcal{O}\!\left(\sum_{\ell=1}^{L} M_\ell\ \cdot C(N,D_\ell)\right)$, where the active model pool $M_\ell$ decreases monotonically with level and periodic feature compression constrains $D_\ell$. Consequently, runtime growth with depth is typically much lower than the naive linear-in-$L$ bound (and may be sub-linear in practice under strong pruning).

\subsection{Stack-of-stacking}

At the end of the recursive pipeline, RocketStack performs a final aggregation step referred to as stack-of-stacking. In this stage, the meta-features generated from each recursive level specifically the outputs that remain after feature selection are accumulated and merged to form a global feature representation. This process results in a single, comprehensive feature matrix composed of the selected, compressed, and pruned outputs across all recursive depths. By aggregating these meta-level representations, the stack-of-stacking layer captures the hierarchical learning dynamics embedded throughout the RocketStack architecture. This final representation is then passed to a final-level estimators (the models survived until level-10), enabling the model to make predictions based on a holistic view of the recursive ensembling trajectory.

\subsection{Statistical testing}

To evaluate the impact of feature selection and model pruning strategies on classification performance, we conducted a series of statistical analyses based on Linear Mixed Models (LMMs). In all models, classification accuracy was treated as the dependent variable, and the experimental factor of interest (e.g., feature-selection strategy or pruning configuration) was specified as a fixed effect. To account for the repeated-measures structure induced by recursive depth, ensembling level (levels 1-10) was included as a random effect to model within-model dependence across levels. First, separate LMMs compared the no-feature-compression baseline against periodic feature-selection variants for both binary and multi-class settings. Next, additional LMMs contrasted each-level feature selection with its corresponding periodic variant across four sub-scenarios: one for binary (SFE each-level vs.\ periodic SFE) and three for multi-class (SFE, autoencoder-based compression, and attention-based selection; each-level vs.\ periodic). Given the multiplicity of comparisons, p-values associated with the fixed-effect terms were adjusted using the Benjamini--Hochberg False Discovery Rate (FDR) procedure. A separate LMM compared three pruning configurations: strict pruning based on unperturbed out-of-fold (OOF) performance ($\lambda=0$), light OOF-score randomization (Gaussian noise; $\lambda=0.05$), and moderate randomization ($\lambda=0.10$), across both classification settings; fixed-effect p-values were again FDR-adjusted. In addition, an LMM was fitted to assess the effect of applying Bayesian HPO at the base-model level relative to the default (untuned) configuration. For LMMs exhibiting significant fixed effects, follow-up pairwise post-hoc comparisons were performed with Holm-adjusted p-values to control the family-wise error rate. All LMMs were evaluated using the Kenward--Roger approximation with Type II sums of squares, and all post-hoc procedures followed Holm's sequential method. Statistical analyses were performed in JASP (v0.16.3.0) \cite{JSSv088i02}.

To examine systematic trends in classification accuracy across increasing ensembling depths, LMMs were fitted separately for each feature selection strategy in both binary and multi-class classification tasks. The models were applied to accuracy scores averaged over folds and datasets at each ensembling level per classifier. Each model included a fixed-effect linear term to capture progression across levels, and classifier identity was treated as a random intercept to account for repeated measurements. Quadratic terms were not included. In total, 15 LMMs were constructed five for binary and ten for multi-class settings. Bonferroni correction was used to adjust for multiple comparisons across strategies. All models employed Type II sums of squares and the Kenward–Roger approximation for degrees of freedom. Python 3.10 with \textit{statsmodels} package was used for the trend specific statistical tests.

\vspace{1cm}

\begin{algorithm}
\setstretch{1.2}
\fontsize{8}{10}\selectfont

\SetAlgoLined
\KwIn{Training data $(X,Y)$; number of levels $L$; feature‐selection flag $FS$; %
      level-\{3,6,9\} switch $L_{369}$; %
      pruning blur $\lambda\in\{0,0.05,0.1\}$; %
      minimum retained models $t_{\min}$}
\KwOut{Fold-wise metrics for each level and final stack-of-stacking layer}

\textbf{Initialise} model pool $\mathcal{M}$ (20 classifiers for binary and 14 classifiers for multi-class classification)\;
\textbf{Initialise} 5-fold stratified CV\;

\ForEach{fold $(X_{\mathrm{tr}},Y_{\mathrm{tr}},X_{\mathrm{te}},Y_{\mathrm{te}})$}{
  \tcp{Train individual baseline models}
  \ForEach{$m\in\mathcal{M}$}{train $m$ on $X_{\mathrm{tr}}$ and store metrics}

  $X^{(0)}\leftarrow X_{\mathrm{tr}}$;\; $X^{(0)}_{\mathrm{te}}\leftarrow X_{\mathrm{te}}$\;
  $\mathcal{M}^{(0)}\leftarrow \mathcal{M}$\;

  initialise global SoS train/test sets: $X_{\text{SoS}}\leftarrow X^{(0)}$, $X_{\text{SoS,te}}\leftarrow X^{(0)}_{\mathrm{te}}$\;

  \For{$\ell = 1$ \KwTo $L$}{
     \tcp{Generate blended train/test sets}
     obtain OOF probabilities $\mathcal{P}^{(\ell)}$ for $\mathcal{M}^{(\ell-1)}$ on $X^{(\ell-1)}$\;
     $X^{(\ell)}\gets [\mathcal{P}^{(\ell)}, X^{(\ell-1)}]$\;
     $X^{(\ell)}_{\mathrm{te}}\gets [m(X^{(\ell-1)}_{\mathrm{te}}), X^{(\ell-1)}_{\mathrm{te}}]$\;

     \tcp{Optional feature selection}
     \If{$FS$ \textbf{and} \big(($L_{369}$ \textbf{and} $\ell\in\{3,6,9\}$) \textbf{or} ($\neg L_{369}$ \textbf{and} $\ell>0$)\big)}{
        $X^{(\ell)}, X^{(\ell)}_{\mathrm{te}} \gets \mathrm{FeatureCompression}(X^{(\ell)}, Y_{\mathrm{tr}})$\;
     }

     \tcp{Meta-learning}
     \ForEach{$m\in\mathcal{M}^{(\ell-1)}$}{train $m$ on $X^{(\ell)}$; evaluate on $X^{(\ell)}_{\mathrm{te}}$}

     \tcp{Blurred pruning}
     compute performance (ROC-AUC for BC and accuracy for MC) vector $\mathbf{a}$ for $\mathcal{M}^{(\ell-1)}$\;
     $\mathbf{a} \gets \mathbf{a} + \mathcal{N}(0, \lambda \cdot \mathrm{range}(\mathbf{a}))$\;
     retain models with $a_i \ge Q_p(\mathbf{a})$, where $Q_p$ is adaptive percentile\;

     \If{retained $< t_{\min}$}{\textbf{break} loop}
     $\mathcal{M}^{(\ell)} \gets$ retained models\;

     \tcp{Update SoS set}
     $X_{\text{SoS}} \gets [X_{\text{SoS}}, X^{(\ell)}]$\;
     $X_{\text{SoS,te}} \gets [X_{\text{SoS,te}}, X^{(\ell)}_{\mathrm{te}}]$\;
  }

  \tcp{Final stack-of-stacking fusion}
  \ForEach{$m\in\mathcal{M}$}{train on $X_{\text{SoS}}$ and evaluate on $X_{\text{SoS,te}}$}
}
\caption{RocketStack architecture}
\label{alg:rocketstack_pseudocode}
\end{algorithm}

\section{Analysis and Results}

\subsection{Systematic evaluation of RocketStack design variants}

To maintain consistency across evaluations, accuracy was used as the primary metric for comparing RocketStack design variants across ensembling depth, while complementary metrics (F1-score, precision, recall, and log-loss) were also reported. Accuracy used as a unified and interpretable baseline for both binary and multi-class settings. To summarize overall performance across RocketStack variants, grand-average metrics were computed over levels 1--10 for both task types. In both settings, the no-compression configuration yielded the highest accuracy overall. Periodic feature compression, where SFE, autoencoders, or attention were applied at levels 3, 6, and 9, generally outperformed each-level compression, with attention-based selection achieving the best average accuracy, followed by periodic autoencoders and SFE. By contrast, continuous compression at every level (for example, SFE each-level) produced the lowest performance across tasks, indicating that selective periodic application is more effective than per-level compression. Finally, light OOF-score randomization during pruning yielded better overall performance than strict OOF pruning in both binary and multi-class settings. Detailed comparisons between variants are provided in the following subsections.

\subsubsection{Comparison between uncompressed fusion and periodic feature selection}

A direct comparison between the no-compression configuration and the corresponding OOF-strict periodic compression variants showed statistically significant accuracy reductions when compression was applied at fixed levels. For methodological consistency, the best-performing periodic setup within each compression family was used as the comparator, namely periodic SFE at levels 3, 6, and 9 for binary classification and periodic Attention at levels 3, 6, and 9 for multi-class classification. Both comparators employed OOF-strict pruning, matching the pruning scheme used in the no-compression setting. In binary classification, no compression achieved 88.52 $\pm$ 1.46\% accuracy versus 88.02 $\pm$ 2.12\% for periodic SFE (see \autoref{tab:grandaverageperformances_of_binaryandmulticlassclassification_merged}), and this gap was supported by a significant difference in estimated marginal means (EMM = 0.885 vs. 0.881; 95\% CI = [0.881, 0.890] vs. [0.876, 0.885]; $p = .019$, FDR-corrected). In multi-class classification, no compression achieved 94.10 $\pm$ 0.94\% versus 93.29 $\pm$ 0.98\% for periodic Attention (see \autoref{tab:grandaverageperformances_of_binaryandmulticlassclassification_merged}), with a similarly significant EMM difference (EMM = 0.941 vs. 0.933; 95\% CI = [0.936, 0.946] vs. [0.928, 0.938]; $p < .001$, FDR-corrected). A visual comparison is provided in \autoref{fig:featureselectioncomparison}B, and the corresponding statistical results are reported in \autoref{tab:lmm_nofs_vs_periodic}.

\subsubsection{Comparison between feature selection frequency (each level and periodically)}

To evaluate the influence of feature selection frequency, comparisons were conducted between each-level feature selection and periodic counterparts (applied at levels 3, 6, and 9) in both binary and multi-class classification. For consistency, only OOF-strict variants were used for periodic strategies, matching each-level feature selection, which also relied exclusively on OOF-strict pruning. Four pairwise LMMs were tested and subjected to a unified FDR correction. In binary classification, periodic SFE (EMM = 0.881, CI = [0.875, 0.886]) significantly outperformed each-level SFE (EMM = 0.869, CI = [0.863, 0.875]; $p = .026$, FDR-corrected; 88.08 $\pm$ 2.12 vs. 86.90 $\pm$ 3.45\% average accuracy). In multi-class classification, periodic SFE (EMM = 0.934, CI = [0.910, 0.959]) similarly outperformed its each-level counterpart (EMM = 0.855, CI = [0.831, 0.879]; $p = .004$, FDR-corrected; 93.45 $\pm$ 1.21\% vs. 85.52 $\pm$ 4.36\% accuracy). A comparable pattern emerged for autoencoders, where periodic application (EMM = 0.899, CI = [0.887, 0.910]) yielded significantly higher performance than the each-level configuration (EMM = 0.878, CI = [0.867, 0.890]; $p = .026$, FDR-corrected; 89.85 $\pm$ 3.61\% vs. 87.84 $\pm$ 3.57\% accuracy). In contrast, attention-based selection showed no significant difference between each-level (EMM = 0.932, CI = [0.928, 0.936]) and periodic variants (EMM = 0.933, CI = [0.929, 0.937]; $p = .600$, FDR-corrected; 93.17 $\pm$ 1.33\% vs. 93.29 $\pm$ 0.98\% accuracy), although a marginal increase of 0.12\% favored periodic selection. Visual differences are shown in \autoref{fig:featureselectioncomparison}C, with corresponding statistical results provided in \autoref{tab:lmm_eachlevel_vs_periodic}.

\subsubsection{Model pruning comparisons: OOF-strict vs. randomized pruning}

\begin{figure}
 \centering \includegraphics[width=0.71\textwidth,keepaspectratio=false]{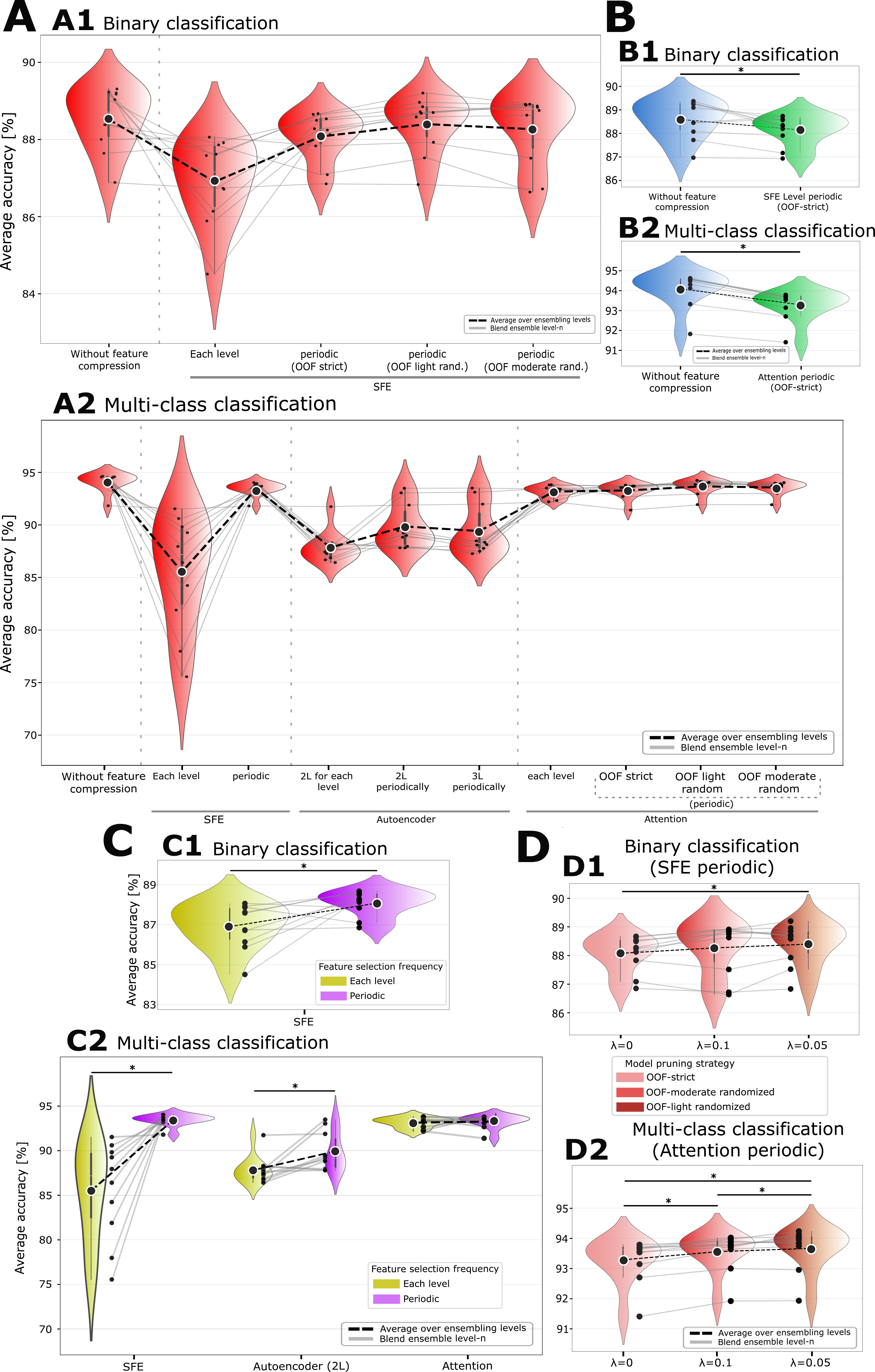}
\caption{Estimation performance across ensembling depths for different feature selection strategies in recursive blend ensembling. Each violin plot shows accuracy distributions across datasets for a given strategy, with the thick black dashed line marking the mean and thin gray lines tracking individual trends. \textbf{A:} Compares overall accuracy (\%) across strategies, including a no-feature-selection baseline, for binary (A1) and multi-class (A2) tasks. Pruning variants—strict OOF-based pruning and randomized OOF score perturbation ($\lambda = 0$: none, $0.05$: light, $0.1$: moderate)—are applied to the SFE strategy (binary) and attention-based strategy (multi-class). \textbf{B:} Highlights baseline vs. best-performing periodic variants—SFE for binary (B1), attention for multi-class (B2)—under strict pruning. \textbf{C:} Assesses selection frequency patterns, contrasting per-level vs. periodic (levels 3, 6, 9) for binary (C1) and multi-class (C2); includes SFE, autoencoder, and attention-based strategies. \textbf{D:} Evaluates RocketStack under OOF-score randomizations prior to pruning: in binary (D1), light randomization outperforms strict pruning; moderate offers no further gain. In multi-class (D2), both light and moderate outperform strict pruning, with light performing best. Asterisks (*) indicate statistically significant differences (FDR-corrected, $p < .05$); 2L: 2-layer autoencoder; 3L: 3-layer autoencoder; \textit{periodic} refers to level 3, 6 and 9.}

  \label{fig:featureselectioncomparison} 
\end{figure}

In an ablation analysis of the pruning mechanism, OOF-strict pruning was compared with randomized variants in which Gaussian noise was injected into OOF performance scores at each level (light randomization, $\lambda = 0.05$; moderate randomization, $\lambda = 0.10$) in both binary and multi-class classification using linear mixed models. All reported $p$-values are Holm-corrected. In binary classification, a significant pruning effect was observed ($F(2, 18) = 5.401$, $p = .015$). Post-hoc tests indicated higher performance for both randomized variants relative to OOF-strict, with light randomization showing a significant improvement (mean diff. = $+0.003$, $z = -3.268$, $p = .003$; 88.08 $\pm$ 2.12\% vs.\ 88.40 $\pm$ 1.72\%) and moderate randomization showing a non-significant trend (mean diff. = $+0.002$, $z = -1.934$, $p = .106$; 88.08 $\pm$ 2.12\% vs.\ 88.27 $\pm$ 1.63\%). No significant difference was observed between the two randomized conditions (mean diff. = $+0.001$, $z = 1.334$, $p = .182$; 88.27 $\pm$ 1.63\% vs.\ 88.40 $\pm$ 1.72\%). In multi-class classification, a robust pruning effect was observed ($F(2, 18) = 26.986$, $p < .001$), with both light and moderate randomization significantly outperforming OOF-strict (mean diff. = $+0.004$, $z = -7.135$, $p < .001$; 93.29 $\pm$ 0.98\% vs.\ 93.67 $\pm$ 1.08\%, and mean diff. = $+0.001$, $z = -5.083$, $p < .001$; 93.29 $\pm$ 0.98\% vs.\ 93.56 $\pm$ 0.97\%, respectively). Light randomization also yielded slightly higher accuracy than moderate randomization (mean diff. = $+0.001$, $z = 2.052$, $p = .040$; 93.67 $\pm$ 1.08\% vs.\ 93.56 $\pm$ 0.97\%). These effects are visualized in \autoref{fig:featureselectioncomparison}D, with corresponding statistical results reported in \autoref{tab:lmm_pruning_oof}.

An additional observation in the multi-class setting was that the stack-of-stack ensemble consistently underperformed, with accuracy never exceeding 93.56\% and elevated log-loss values in some cases (e.g., 0.73 $\rightarrow$ 1.34), supporting that progressive level-wise ensembling is more effective than final-stage aggregation alone. Summary tables are provided in \autoref{tab:grandaverageperformances_of_binaryandmulticlassclassification_merged}, an overall visual summary is shown in \autoref{fig:featureselectioncomparison}A (A1--A2), and dataset-wise results are reported in Supplementary \textbf{\textit{Figure S1}} (binary) and \textbf{\textit{Figure S2}} (multi-class).


\begin{table}[H]
\fontsize{6}{7}\selectfont
\centering
\caption{Grand average performance metrics across RocketStack ensembling depths (levels 1–10), comparing different RocketStack variants in binary and multi-class classification. Values represent means and standard deviations aggregated over all models, datasets, and 5-fold cross-validation splits. The label \textit{periodic} refers to feature selection applied specifically at levels 3, 6, and 9.}
\label{tab:grandaverageperformances_of_binaryandmulticlassclassification_merged}

\begin{tabular}{lllllll}
\toprule
Classification type & Feature selection & Accuracy [\%] & F1-score [\%] & Precision [\%] & Recall [\%] & Logloss \\
\midrule
Binary & Without feature compression & 88.52 ± 1.46 & 87.68 ± 1.41 & 88.17 ± 1.46 & 88.52 ± 1.46 & 0.67 ± 0.14 \\
Binary & SFE each level & 86.90 ± 3.45 & 85.37 ± 3.64 & 85.65 ± 3.36 & 86.90 ± 3.45 & 0.59 ± 0.20 \\
Binary & SFE periodic (OOF-strict) & 88.08 ± 2.12 & 87.09 ± 2.23 & 87.50 ± 2.11 & 88.08 ± 2.12 & 0.64 ± 0.19 \\
Binary & SFE periodic (OOF-light rand.) & 88.40 ± 1.72 & 87.34 ± 1.75 & 87.70 ± 1.87 & 88.40 ± 1.72 & 0.64 ± 0.16 \\
Binary & SFE periodic (OOF-moderate rand.) & 88.27 ± 1.63 & 87.18 ± 1.88 & 87.38 ± 2.25 & 88.27 ± 1.63 & 0.68 ± 0.18 \\
\midrule
Multi-class & Without feature compression & 94.10 ± 0.94 & 93.98 ± 0.96 & 94.13 ± 0.95 & 94.10 ± 0.94 & 0.42 ± 0.09 \\
Multi-class & SFE each level & 85.52 ± 4.36 & 84.23 ± 5.09 & 85.00 ± 4.89 & 85.52 ± 4.36 & 1.08 ± 0.41 \\
Multi-class & SFE periodic (OOF-strict) & 93.45 ± 1.21 & 93.31 ± 1.29 & 93.52 ± 1.26 & 93.45 ± 1.21 & 0.44 ± 0.10 \\
Multi-class & Autoencoders each level, 2L & 87.84 ± 3.57 & 86.92 ± 3.94 & 87.30 ± 4.01 & 87.84 ± 3.57 & 1.16 ± 0.54 \\
Multi-class & Autoencoders periodic, 2L & 89.85 ± 3.61 & 89.20 ± 4.02 & 89.88 ± 3.99 & 89.85 ± 3.61 & 0.77 ± 0.34 \\
Multi-class & Autoencoders periodic, 3L & 89.41 ± 3.68 & 88.63 ± 4.18 & 89.04 ± 4.20 & 89.41 ± 3.68 & 0.96 ± 0.38 \\
Multi-class & Attention each level & 93.17 ± 1.33 & 92.99 ± 1.43 & 93.26 ± 1.38 & 93.17 ± 1.33 & 0.46 ± 0.12 \\
Multi-class & Attention periodic (OOF-strict) & 93.29 ± 0.98 & 93.20 ± 1.00 & 93.43 ± 0.99 & 93.29 ± 0.98 & 0.45 ± 0.09 \\
Multi-class & Attention periodic (OOF-light rand.) & 93.67 ± 1.08 & 93.53 ± 1.11 & 93.77 ± 1.06 & 93.67 ± 1.08 & 0.45 ± 0.10 \\
Multi-class & Attention periodic (OOF-moderate rand.) & 93.56 ± 0.97 & 93.47 ± 0.99 & 93.67 ± 0.99 & 93.56 ± 0.97 & 0.51 ± 0.09 \\
\bottomrule
\end{tabular}
\end{table}

\begin{table}[H]
\fontsize{7}{8}\selectfont
\centering
\caption{Statistical comparison of feature selection strategies and model pruning conditions on binary and multi-class settings.}
\label{tab:statisticaltest_featureselectionstrategies_modelpruning}

\begin{subtable}[t]{\textwidth}
\centering
\caption{LMM with random effect grouping factor of each ensembling level between \emph{Without feature compression} and \emph{Periodic SFE} in RocketStack for both binary and multi-class settings.}
\label{tab:lmm_nofs_vs_periodic}

\begin{tabular}{llccccc}
\toprule
\multirow{2}{*}{\textbf{Classification}} & \multirow{2}{*}{\textbf{Condition}} & \multicolumn{2}{c}{\textbf{\emph{p}-value}} & \multicolumn{2}{c}{\textbf{95\% CI}} & \textbf{EMM} \\
\cmidrule(lr){3-4} \cmidrule(lr){5-6}
& & Uncorr. & FDR-corr. & Lower & Upper & \\
\midrule
\multirow{2}{*}{Binary} 
& Without feature compression & \multirow{2}{*}{.019} & \multirow{2}{*}{.019} & 0.881 & 0.890 & 0.885 \\
& Periodic SFE (L3/6/9) & & & 0.876 & 0.885 & 0.881 \\
\midrule
\multirow{2}{*}{Multi-class} 
& Without feature compression & \multirow{2}{*}{$<.001$} & \multirow{2}{*}{$<.001$} & 0.936 & 0.946 & 0.941 \\
& Periodic Attention (L3/6/9) & & & 0.928 & 0.938 & 0.933 \\
\bottomrule
\end{tabular}
\end{subtable}

\vskip 2mm

\begin{subtable}[t]{\textwidth}
\centering
\caption{LMM with random effect grouping factor of each ensembling level between feature selection in each level and periodic in RocketStack for both binary and multi-class settings.}
\label{tab:lmm_eachlevel_vs_periodic}

\begin{tabular}{llccccc}
\toprule
\multirow{2}{*}{\textbf{Classification}} & \multirow{2}{*}{\textbf{Condition}} & \multicolumn{2}{c}{\textbf{\emph{p}-value}} & \multicolumn{2}{c}{\textbf{95\% CI}} & \textbf{EMM} \\
\cmidrule(lr){3-4} \cmidrule(lr){5-6}
& & Uncorr. & FDR-corr. & Lower & Upper & \\

\midrule
\multirow{2}{*}{Binary} 
& Each level SFE & \multirow{2}{*}{.019} & \multirow{2}{*}{.026} & 0.863 & 0.875 & 0.869 \\
& Periodic SFE (L3/6/9) & & & 0.875 & 0.886 & 0.881 \\

\midrule
\multirow{2}{*}{Multi-class} 
& Each level SFE & \multirow{2}{*}{.001} & \multirow{2}{*}{.004} & 0.831 & 0.879 & 0.855 \\
& Periodic SFE (L3/6/9) & & & 0.910 & 0.959 & 0.934 \\

\midrule
\multirow{2}{*}{Multi-class} 
& Each level Autoencoders & \multirow{2}{*}{.020} & \multirow{2}{*}{.026} & 0.867 & 0.890 & 0.878 \\
& Periodic Autoencoders (L3/6/9) & & & 0.887 & 0.910 & 0.899 \\

\midrule
\multirow{2}{*}{Multi-class} 
& Each level Attention & \multirow{2}{*}{.600} & \multirow{2}{*}{.600} & 0.928 & 0.936 & 0.932 \\
& Periodic Attention (L3/6/9) & & & 0.929 & 0.937 & 0.933 \\

\bottomrule
\end{tabular}
\end{subtable}

\vskip 2mm


\begin{subtable}[t]{\textwidth}
\centering
\caption{LMM results comparing model pruning scenarios (OOF-strict vs. randomized) in RocketStack for binary and multi-class settings.}
\label{tab:lmm_pruning_oof}

\begin{tabularx}{\textwidth}{X X}
\begin{minipage}[t]{\linewidth}
\textbf{  (c1) Binary classification} \\[1ex]
\begin{tabular}{lccc}
\toprule
df & F & \multicolumn{2}{c}{\textbf{\emph{p}-value}} \\
\cmidrule(lr){3-4}
& & Uncorr. & FDR-corr. \\
\midrule
2, 18 & 5.401 & .015 & .015 \\
\bottomrule
\end{tabular}

\vskip 1mm

\begin{tabular}{llccl}
Group 1 & Group 2 & Mean diff. & z & $P_{Holm}$ \\
\midrule
No rand. & Rand-0.05 & -0.003 & -3.268 & .003 \\
No rand. & Rand-0.1 & -0.002 & -1.934 & .106 \\
Rand-0.05 & Rand-0.1 & -0.001 & 1.334 & .182 \\
\bottomrule
\end{tabular}
\end{minipage}
&
\begin{minipage}[t]{\linewidth}
\textbf{  (c2) Multi-class classification} \\[1ex]
\begin{tabular}{lccc}
\toprule
df & F & \multicolumn{2}{c}{\textbf{\emph{p}-value}} \\
\cmidrule(lr){3-4}
& & Uncorr. & FDR-corr. \\
\midrule
2, 18 & 26.986 & $<.001$ & $<.001$ \\
\bottomrule
\end{tabular}

\vskip 1mm

\begin{tabular}{llccl}
Group 1 & Group 2 & Mean diff. & z & $P_{Holm}$ \\
\midrule
No rand. & Rand-0.05 & -0.004 & -7.135 & $<.001$ \\
No rand. & Rand-0.1 & -0.001 & -5.083 & $<.001$ \\
Rand-0.05 & Rand-0.1 & 0.001 & 2.052 & .040 \\
\bottomrule
\end{tabular}
\end{minipage}
\end{tabularx}
\end{subtable}


\end{table}

\subsubsection{Trend analysis of performance across ensembling depths}

Systematic trends in accuracy across ensembling depths were observed in both binary and multi-class settings. All statistical results reported here reflect Bonferroni-corrected p-values. To ensure valid and comparable trend analysis across ensembling levels, an additional filtering step was applied: only models that remained present at all levels in at least three datasets were included. This measure was taken to avoid bias introduced by models that appeared in only a small number of datasets, which could otherwise disproportionately influence performance estimates. In the binary classification domain, statistically significant trends were found in most configurations, including the baseline without feature compression ($p < .001$), periodic SFE with OOF-strict ($p = .004$), and both randomized variants (light and moderate; $p < .001$ for both). These configurations corresponded to mean accuracy improvements of $6.15\%$, $4.92\%$, $5.24\%$, and $5.20\%$ (see \autoref{fig:trendanalysisresults_on_ensembling_depth}) from individual models to RocketStack level-10, respectively. In contrast, the SFE each-level strategy showed no statistically significant trend ($p = .411$) and only a marginal gain of $0.20\%$. In the multi-class setting, the baseline without feature compression ($p < .001$) and both SFE strategies at each-level ($p < .001$) and periodic ($p = .013$) exhibited significant trends, with respective accuracy changes of $13.72\%$, $-7.90\%$, and $12.69\%$. While periodic SFE yielded substantial gains, the each-level variant showed a negative mean change despite statistical significance, likely due to variability across datasets. Attention-based feature selection demonstrated statistically significant trends in all periodic configurations: OOF-strict ($p = .002$), OOF-light ($p = .003$), and OOF-moderate ($p = .026$), with corresponding accuracy improvements of $12.88\%$, $13.94\%$, and $13.38\%$. The Attention each-level variant did not show statistical significance ($p = .087$) but still achieved an average gain of $11.60\%$. In contrast, none of the autoencoder-based strategies produced significant trends (all $p = 1.000$), with only moderate performance increases ranging from $4.40\%$ to $6.90\%$. These findings emphasize that periodic feature selection particularly Attention-based and OOF-randomized configurations not only drives statistically detectable trends but also delivers substantial accuracy improvements across RocketStack ensembling depths, whereas autoencoder-based and each-level SFE strategies lack consistent benefits. Summary statistics from ensembling depth analysis for binary and multi-class classification are provided in \autoref{tab:averageperformances_of_binaryclassification_merged} and \autoref{tab:averageperformances_of_multiclassclassification_merged}. The statistical results of the trend analyses are presented in \autoref{tab:trend_analysis}. Trend visualizations across binary and multi-class settings including performance trajectories, accuracy deltas, and classifier-level performance evolutions heatmaps over recursive stacking are presented in \autoref{fig:trendanalysisresults_on_ensembling_depth}.

\subsection{Benchmarking RocketStack against top performing ensemble performers}

\begin{figure}
 \centering \includegraphics[width=0.81\textwidth,keepaspectratio=false]{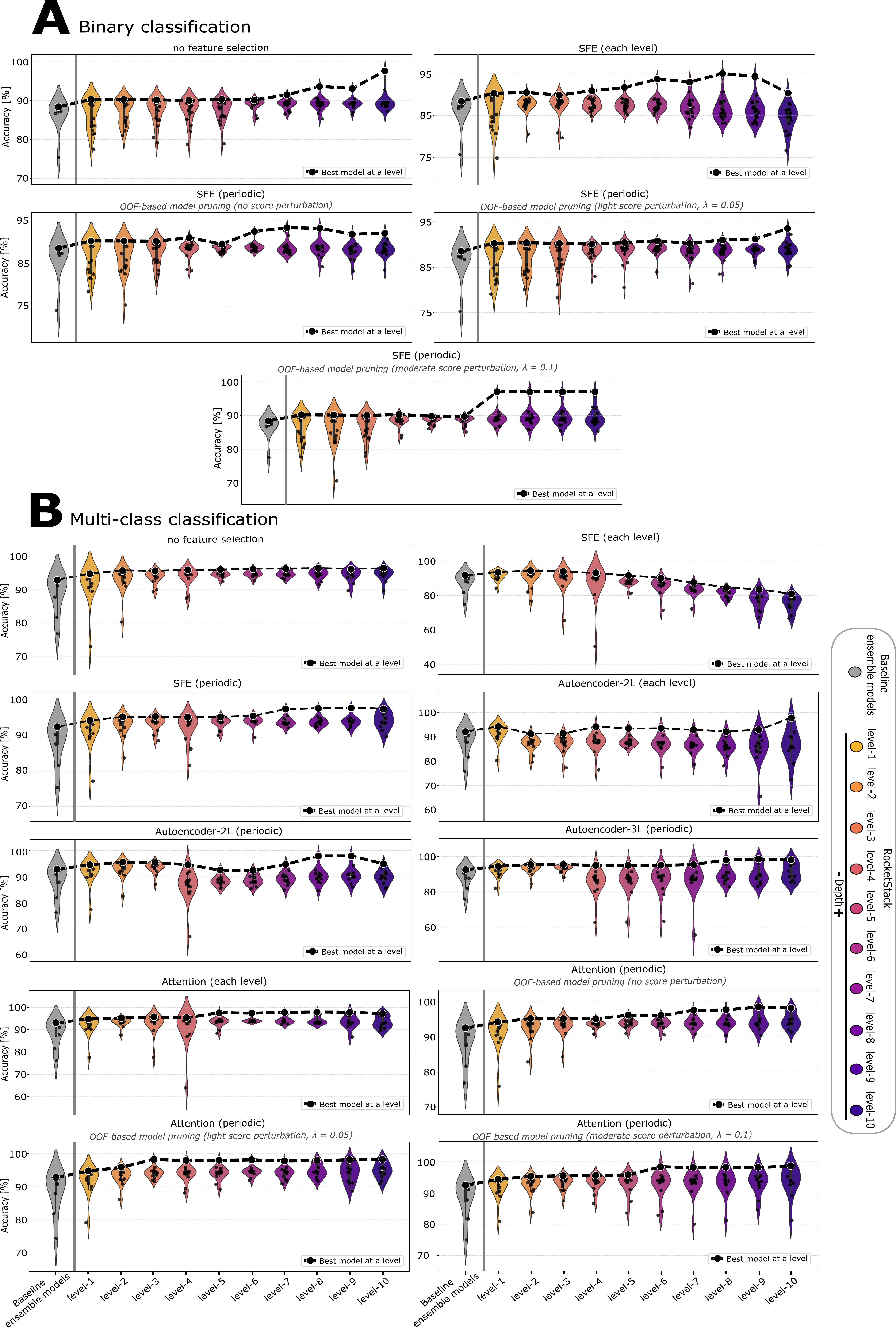}
 \caption{Performance comparison of RocketStack across ensembling levels against top-performing baseline ensemble models in binary \textbf{(A)} and multi-class \textbf{(B)} classification tasks. Each violin plot represents the distribution of accuracy scores across different ensembling levels, with individual baseline ensemble models shown in gray. The dashed black line connects the highest-performing model at each ensembling level, highlighting performance trends as depth increases. Different feature selection strategies (SFE, autoencoder-based, attention-based) are evaluated to assess their impact on ensembling effectiveness. \textit{periodic} refers to feature selection applied specifically at levels 3, 6, and 9. Overall, deeper RocketStack levels consistently yield superior or comparable performance to baseline ensemble models across various scenarios.} 
  \label{fig:bestvsbest_benchmark} 
\end{figure}

To benchmark RocketStack against standard ensemble methods, performance was compared between selected level-0 baselines and the meta-classifiers used across levels 1 to 10. For both binary and multi-class classification, the baseline pool included \textit{xgb}, \textit{lgbm}, \textit{randforest}, \textit{bagging}, \textit{adaboost}, \textit{extra trees}, \textit{hist gradient boosting}, \textit{catboost}, and \textit{ngboost}. Notably, \textit{ngboost} was included only as a benchmark and was not part of RocketStack's model pool. The same filtering criterion used in the trend analysis was applied, such that classifiers not present in at least three datasets at a given level were excluded to ensure consistent comparisons. After filtering, results were averaged over folds and datasets for each classifier at each level, and the best-performing model was selected as the representative baseline at level 0 and as the representative RocketStack meta-classifier at each ensembling level (see \autoref{tab:bestvsbest} and \autoref{fig:bestvsbest_benchmark}).

In binary classification, all RocketStack configurations exceeded the baseline ensemble accuracy of 88.46\% across depth, with distinct trajectories. The no-compression variant showed moderate early gains followed by a pronounced rise after level 6, reaching 97.69\% at level 10. The each-level SFE configuration increased steadily and peaked at level 9 (94.41\%), but declined at level 10 (90.41\%), while remaining above baseline. For periodic SFE, all three pruning modes yielded stable improvements over depth. Relative to deterministic OOF-strict (91.94\% at level 10), randomized pruning produced smoother progressions, with OOF-light reaching 93.54\% at level 10 and OOF-moderate attaining the highest accuracy (97.08\%), although performance plateaued after level 7. The violin plots further illustrate widening margins between RocketStack and baseline models at later levels, consistent with progressive gains from deeper ensembling (\autoref{fig:bestvsbest_benchmark}).

In multi-class classification, all RocketStack configurations except each-level SFE surpassed the baseline ensemble accuracy of 92.49\% by level 10. The no-compression variant improved rapidly early on, reaching 95.99\% at level 6 and plateauing thereafter, with 96.36\% at level 10. Periodic SFE (OOF-strict) increased steadily to level 7 (95.46\%) and continued to improve through level 10 (97.76\%), whereas each-level SFE declined markedly from 93.62\% at level 1 to 80.84\% at level 10. Autoencoder-based compression showed mixed but generally positive trends: the periodic 3-layer variant rose from 94.50\% at level 1 to 98.16\% at level 10, periodic 2-layer peaked at 94.78\% at level 10 with mild volatility, and the each-level 2-layer variant exhibited irregular progression but ended above baseline (97.86\%). Attention-based strategies yielded the most consistent gains. Each-level attention increased from 94.88\% to 97.26\%, while periodic attention variants remained high across depth, with OOF-light peaking at level 3 (98.12\%) and maintaining 98.14\% at level 10. Both OOF-strict and OOF-moderate continued to improve through level 10, reaching 98.19\% and 98.60\%, respectively, with the latter representing the highest accuracy observed in the multi-class benchmark (\autoref{tab:bestvsbest}). These patterns are also reflected in the violin plots, which show stable distributions and upward shifts across levels, particularly for attention-based configurations (\autoref{fig:bestvsbest_benchmark}).

\begin{center}
\includegraphics[width=0.83\textwidth,keepaspectratio=false]{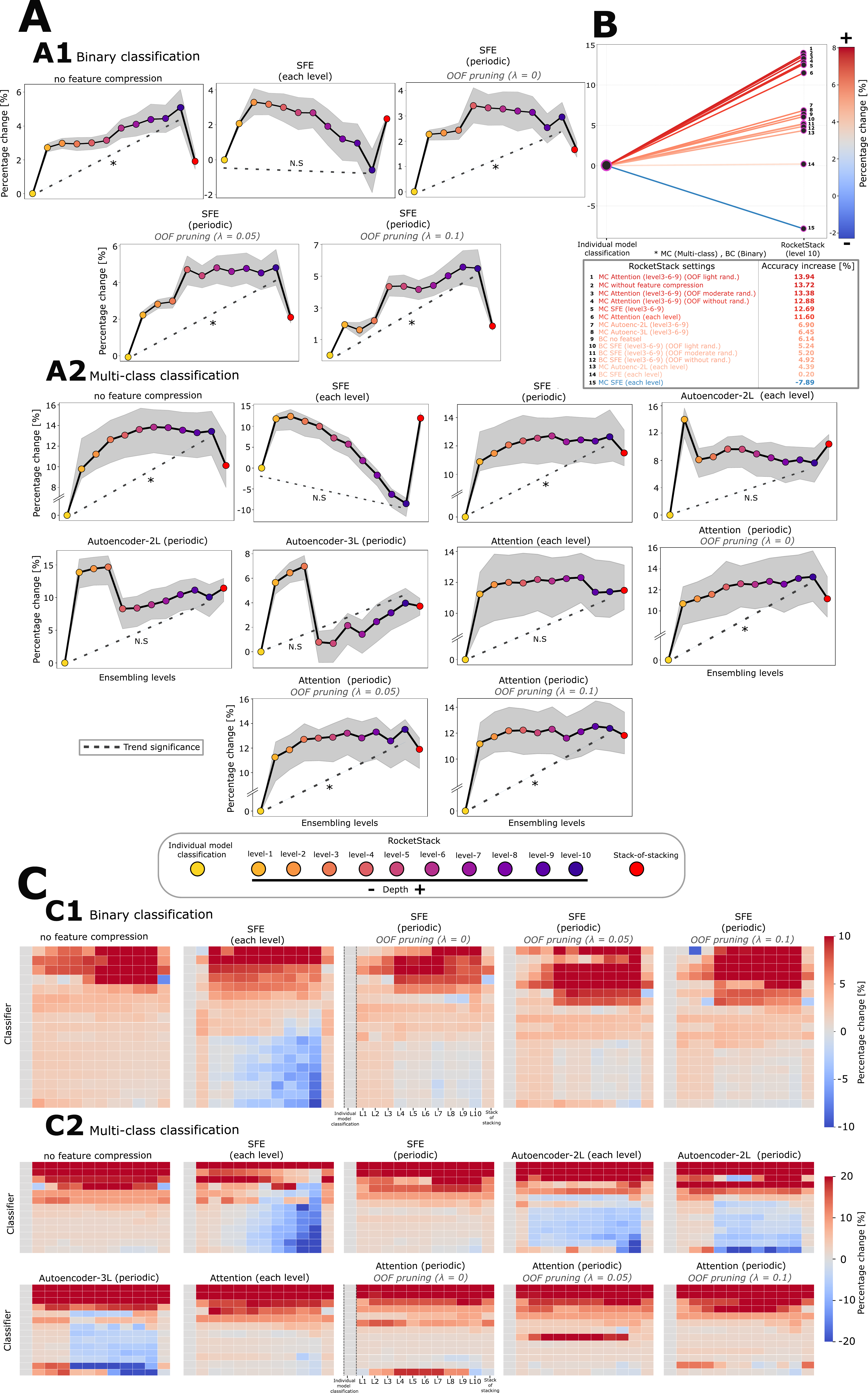}
\captionof{figure}{Trend analysis results across ensembling depth. \textbf{A:} Shows accuracy (\%) trends from baseline (individual) model performance to recursive stacking levels (L1–L10), including stack-of-stacking as the final level. Averages are computed across all surviving models, datasets, and 5-fold splits for both binary (A1) and multi-class (A2) settings. \textit{Broken axis} is used in some multi-class trend visuals (A2) to compress the large level-0 to level-1 gain and better emphasize gradual improvements from level-1 to level-10. Statistical significance of trends was assessed using LMM, comparing performance from the baseline through L10 (excluding stack-of-stacking); asterisks (*) indicate significant trends ($p < .001$), while “N.S.” denotes non-significant results. \textbf{B:} Reports direct accuracy differences between baseline and level-10 performance across all RocketStack variants for binary and multi-class tasks, highlighting the superiority of periodic attention-based feature compression with light OOF-score randomization for pruning. \textbf{C:} Visualizes model-wise performance trajectories from individual models through stack-of-stacking without averaging, using heatmaps to depict progression across ensembling levels. \textit{periodic} refers to level 3, 6 and 9.}
\label{fig:trendanalysisresults_on_ensembling_depth}
\end{center}

\subsection{Runtime progression over levels and comparison across feature selection methods}

To evaluate computational scalability, runtime progression across recursive ensembling levels was analyzed under the evaluated feature selection and model pruning configurations. Elapsed time was recorded per model prediction, per fold, and per dataset, then averaged across folds, datasets, and classifiers and normalized to the $[0,1]$ scale for cross-setting comparison. This analysis is central because deep recursive stacking is often associated with computational inflation as depth and feature dimensionality increase.

In binary classification, runtime growth remained moderate across all configurations. As reported in \autoref{tab:runtimecomparisons}, the no-compression setting increased from 0.167 at level 1 to 0.344 at level 10. Applying SFE at each level produced the steepest growth, reaching 0.389 at level 10, consistent with repeated feature transformations. Periodic SFE was the most runtime-efficient, increasing from 0.167 to 0.308 across levels. This scaling is consistent with the binary setting, where each model contributes a single probabilistic feature, and with consistent model pruning that limits feature expansion.

\begin{figure}
 \centering \includegraphics[width=0.99\textwidth,keepaspectratio=false]{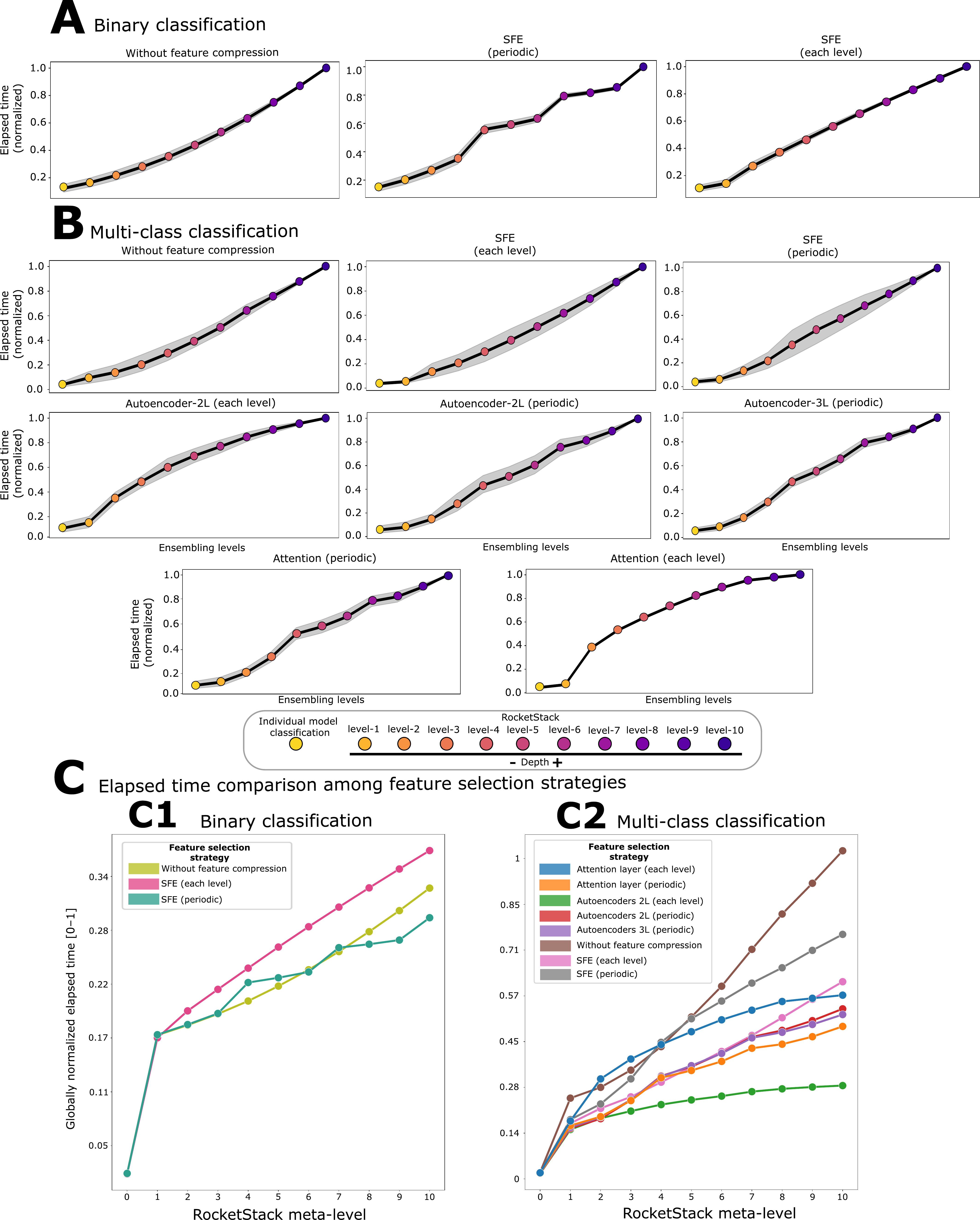}
\caption{Elapsed time progression from baseline classification (level 0) to level 10 across RocketStack configurations for binary (\textbf{A}) and multi-class (\textbf{B}) settings. Each individual subfigure is independently normalized to the [0–1] range. \textbf{C:} Presents a single globally normalized (0–1) runtime trajectory for all feature selection strategies in both binary (\textbf{C1}) and multi-class (\textbf{C2}) tasks, enabling direct comparison of computational efficiency across depth. 2L: 2-layer autoencoder; 3L: 3-layer autoencoder; \textit{periodic} refers to feature selection applied specifically at levels 3, 6, and 9.}
   
  \label{fig:elapsedtimes} 
\end{figure}

In multi-class classification, where each model emits multiple probabilistic outputs, runtime growth was larger. Without feature compression, normalized runtime increased from 0.124 at level 1 to 1.000 at level 10. Each-level SFE moderated this increase (0.593 at level 10), while periodic SFE showed intermediate scaling (0.754). Autoencoders and attention mechanisms yielded more favorable profiles. The 2-layer autoencoder applied at each level showed a reverse-asymptotic trajectory, plateauing after level 5 and reaching 0.237 at level 10. Periodic attention maintained 0.439 at level 10, while each-level attention ended at 0.543 (see \autoref{tab:runtimecomparisons} and \autoref{fig:elapsedtimes}). These trends indicate that pruning and feature compression can counteract the computational demands typically associated with deep ensembling. Since randomized pruning variants do not change the number of models retained per level, only which models are selected, their runtime profiles are identical to their non-randomized counterparts and are omitted from elapsed time comparisons.

\subsection{Number of features and models over leveling}

\begin{figure}
 \centering \includegraphics[width=0.89\textwidth,keepaspectratio=false]{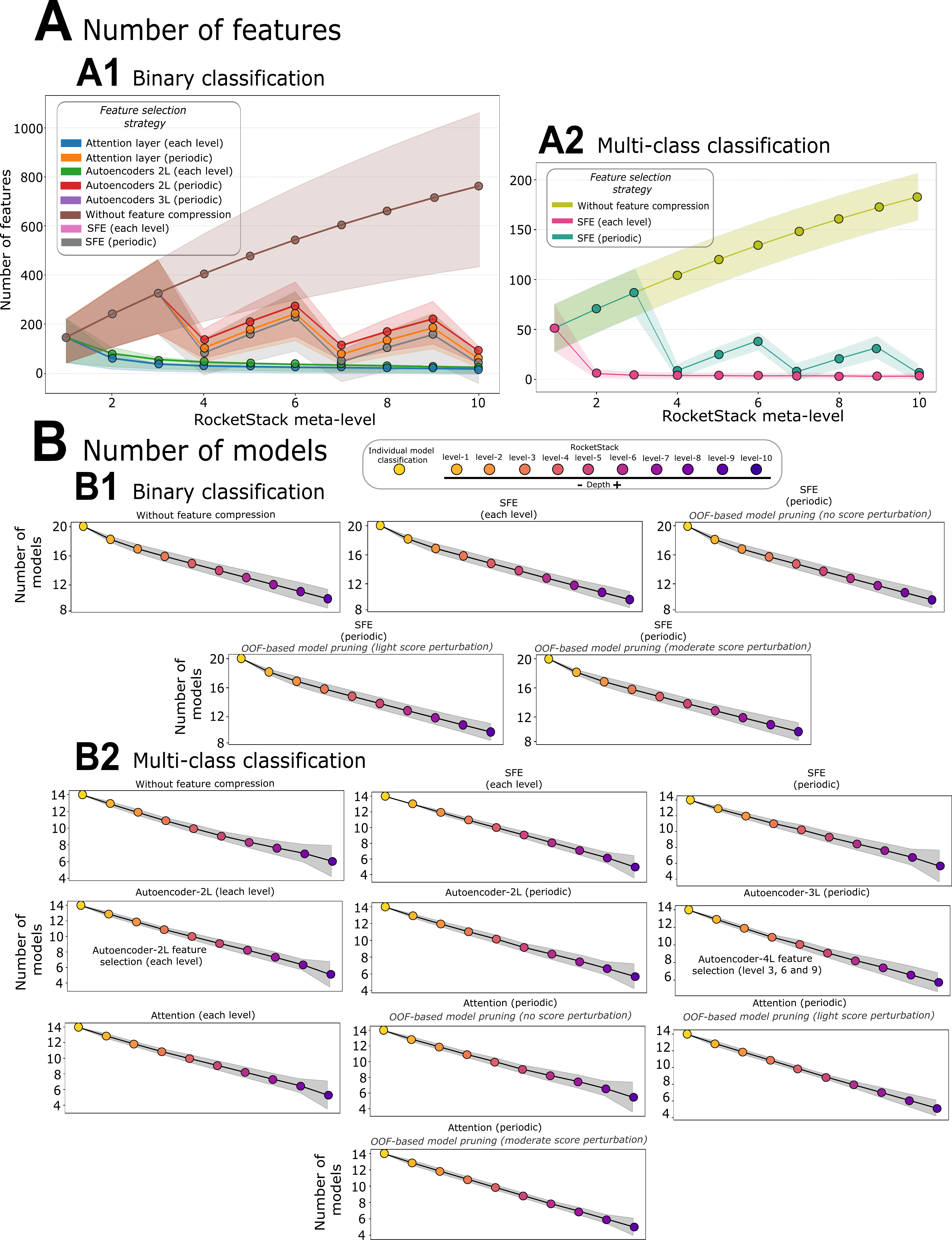}
\caption{Feature and model count progression across RocketStack ensembling levels. \textbf{A:} Shows the number of features retained from baseline to level 10 in binary (\textbf{A1}) and multi-class (\textbf{A2}) settings under various feature selection strategies. In the absence of feature selection, the number of features grows with depth but follows an asymptotically slowing trend—an emergent behavior driven by RocketStack’s pruning mechanism, which limits feature accumulation by progressively eliminating weaker models. In contrast, level-wise and periodic feature selection methods (e.g., SFE, autoencoders, and attention layers) maintain tight control over feature dimensionality, with periodic strategies showing particularly stable compression across depth. \textbf{B:} Illustrates the average number of models surviving at each meta-level for binary (\textbf{B1}) and multi-class (\textbf{B2}) tasks. \textit{periodic} refers to feature selection applied specifically at levels 3, 6, and 9.}
 
  \label{fig:numberoffeatures_over_leveling} 
\end{figure}

The evolution of feature dimensionality across RocketStack ensembling depths was examined to assess how different feature selection strategies control feature inflation. In both binary and multi-class settings, the number of features was averaged per level across all classifiers and datasets, with results summarized in \autoref{tab:number_of_features} and visualized in \autoref{fig:numberoffeatures_over_leveling} A1--A2. In the binary setting, the no-compression configuration exhibited steady feature accumulation from level 1 (51.57) to level 10 (177.30), reflecting stacking without explicit dimensionality control. Since these values are averaged across datasets, feature counts appear as non-integer quantities. Applying SFE at every level produced a consistent and sharp reduction, decreasing from 51.57 at level 1 to 3.05 at level 10. Periodic SFE showed a cyclic accumulation and reduction pattern: the feature count increased between selection checkpoints (e.g., level 1 to 3: 51.57 $\rightarrow$ 86.66) and then dropped sharply after selection (e.g., level 3 to 4: 86.66 $\rightarrow$ 7.55), with subsequent growth phases diminishing due to cumulative pruning. These dynamics illustrate that periodic selection allows controlled recovery of meta-features, whereas per-level selection maintains minimal dimensionality throughout.

In the multi-class setting, feature accumulation was higher and more volatile due to class-probability expansion in OOF-based meta-features. Without compression, the number of features increased from 145.56 at level 1 to 762.91 at level 10, with a visible deceleration in growth rate. Each-level SFE again enforced strict trimming, reducing the feature count to 3.12 at level 10. All periodic compression strategies (SFE, autoencoders, and attention) showed a similar three-level cycle, characterized by feature buildup followed by dimensional reduction. For example, periodic attention rose to 331.24 at level 3 and dropped to 103.42 at level 4, with analogous oscillations continuing through level 10. Among all configurations, each-level attention and each-level 2-layer autoencoders yielded the lowest final dimensionalities, reaching 8.98 and 13.26 features at level 10, respectively. Complete level-wise feature counts for all settings are reported in \autoref{tab:number_of_features}.

Across both settings, the number of models decreased steadily with successive levels due to the built-in pruning mechanism. As shown in \autoref{fig:numberoffeatures_over_leveling} B1--B2, this reduction was approximately linear and largely independent of the feature selection strategy. In binary classification, the model pool decreased from 20 to approximately 8 by level 10, while in multi-class classification it decreased from 14 to approximately 4--6 models by the final level. This consistent pruning behavior supports computational tractability and scalable deployment across depth.

\subsection{Performance comparison of base-level parameter optimization on RocketStack}

\begin{figure}
 \centering \includegraphics[width=1\textwidth,keepaspectratio=false]{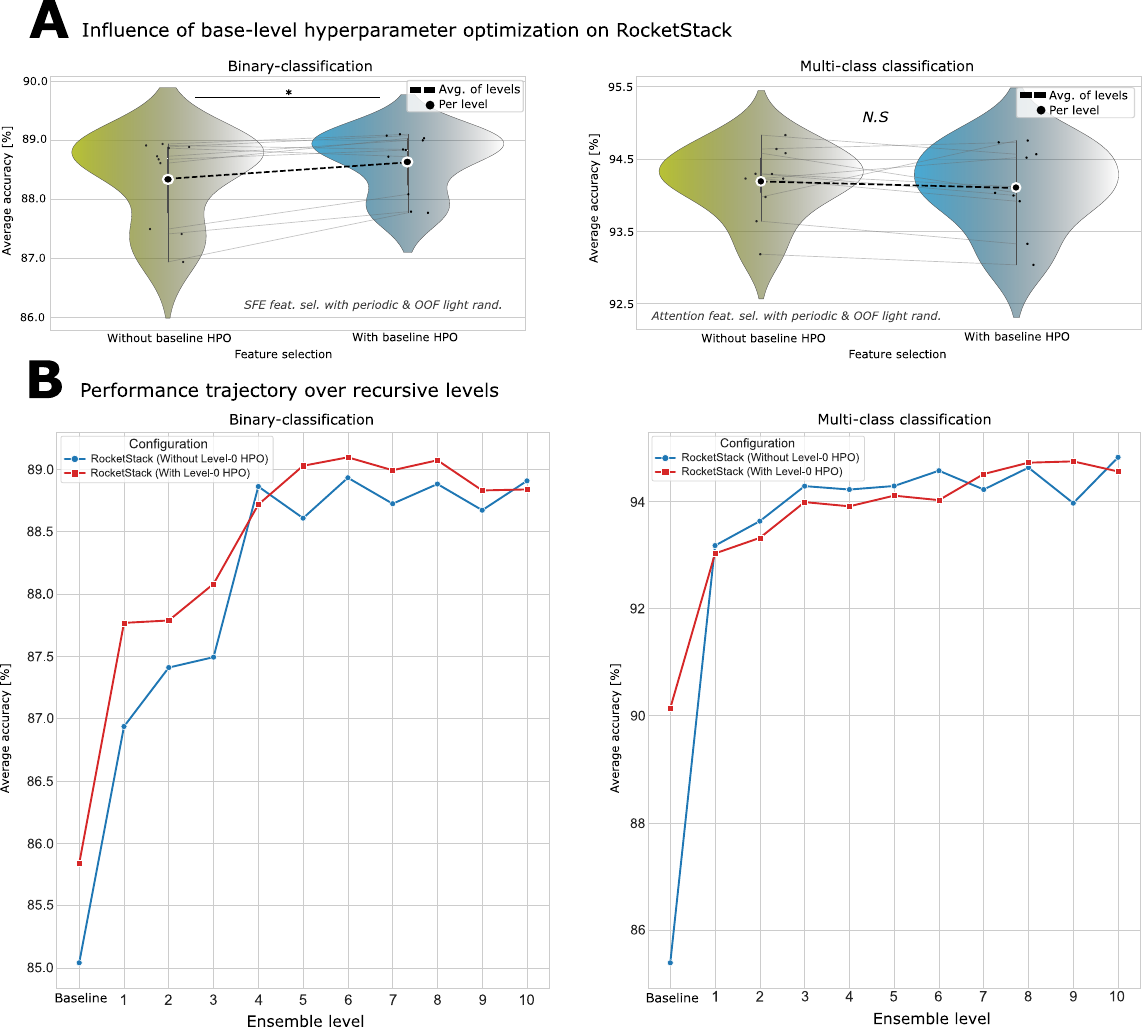}
\caption{Influence of base-level parameter optimization on selected RocketStack setting (incorporating OOF-light randomized model pruning paired with periodic SFE for binary classifications, and periodic Attention feature selection for multi-class classifications).\textbf{(A)} Distribution of average accuracy [\%] for binary-classification and multi-class classification, comparing architectures built without baseline HPO versus those with baseline HPO. \textbf{(B)} The performance trajectory over recursive levels, mapping the average accuracy from the initial baseline models through ensemble level 10. The trajectories contrast the configuration of RocketStack without Level-0 HPO against RocketStack with Level-0 HPO.}
\label{fig:HPO_influence_on_RocketStack} 
\end{figure}

As illustrated in Figure \ref{fig:HPO_influence_on_RocketStack}, applying Bayesian HPO to Level-0 base learners establishes a higher initial accuracy compared to using default parameters. This comparison evaluates the selected top-performing RocketStack configurations at Level-10 specifically, periodic SFE with light randomization for binary classification and periodic Attention with light randomization for multi-class classification. In the binary classification setting, the optimized configuration starts with a performance advantage at Level-0, but the default configuration steadily closes this gap across subsequent recursive levels. By Level-10, the default RocketStack configuration achieves a final accuracy of 88.90\%, slightly outperforming the Level-0 HPO configuration's 88.83\%. A similar trajectory is observed in the multi-class setting, where the HPO configuration begins with a substantial initial lead at Level-0 (90.14\% versus 85.38\%). However, the default configuration ultimately overtakes the optimized setup at deeper levels, culminating in a Level-10 accuracy of 94.82\% compared to 94.56\% for the HPO variant. In the binary setting, base learners optimized with HPO (EMM = 0.886, CI = [0.882, 0.980]) significantly outperformed the un-tuned configuration (EMM = 0.883, CI = [0.879, 0.887]; FDR-corrected p = .03). However, in the multi-class setting, applying HPO to base learners did not yield a significant performance gain; rather, it led to a minor and statistically non-significant overall performance decay (94.09\% average accuracy; EMM = 0.941, CI = [0.938, 0.944]) compared to the default configuration (94.18\% average accuracy; EMM = 0.942, CI = [0.939, 0.945]; F(1,9)=0.553, FDR-corrected p = .476).

Same selected RocketStack configurations at Level-10 for binary and multi-class classifications were also benchmarked against state-of-the-art (SOTA) deep tabular models, TabNet \cite{arik2021tabnet} and Deep Forest \cite{deepforest}, with results summarized in Table \ref{tab:RocketStackvsSOTA}. In binary classification, the default RocketStack configuration (88.90\%) outperforms both Deep Forest (88.39\%) and TabNet (84.71\%), when both baseline models utilizing Bayesian HPO. In the multi-class setting, the default RocketStack configuration achieves 94.82\%, similarly exceeding the optimized Deep Forest (93.53\%) and TabNet (86.55\%) models. Furthermore, while the RocketStack variants incorporating level-0 HPO yield slightly lower accuracy than their default counterparts at level-10, they still surpass all external baselines in both binary (88.83\%) and multi-class (94.56\%) evaluations.

\begin{table}[H]
\fontsize{7}{7}\selectfont
\centering
\caption{Benchmarking RocketStack (Level-10) against optimized SOTA deep tabular baselines} 
\label{tab:RocketStackvsSOTA}

\begin{tabular}{llc@{\hskip 4pt}*{10}{@{\hskip 4pt}c@{\hskip 4pt}}}
\toprule
\textbf{Classification} & \textbf{Model} & \textbf{HPO} & \textbf{Accuracy [\%]} \\

\midrule
Binary & TabNet & Bayesian HPO & 84.71 \\
Binary & Deep Forest & Bayesian HPO & 88.39\\
Binary & RocketStack-Level10 (Periodic SFE (Rand-$\lambda=0.05$)) & Without HPO & 88.90\\
Binary & RocketStack-Level10 (Periodic SFE (Rand-$\lambda=0.05$)) & Bayesian HPO in Level0 & 88.83\\
\midrule
Multi-class & TabNet & Bayesian HPO & 86.55 \\
Multi-class & Deep Forest & Bayesian HPO & 93.53\\
Multi-class & RocketStack-Level10 (Periodic Attention (Rand-$\lambda=0.05$)) & Without HPO & 94.82\\
Multi-class & RocketStack-Level10 (Periodic Attention (Rand-$\lambda=0.05$)) & Bayesian HPO in Level0 & 94.56\\

\bottomrule

\end{tabular}
\end{table}

\section{Discussion}

This study introduced RocketStack, a level-aware recursive ensembling architecture that reimagines deep stacking as a scalable, modular, and computationally efficient learning paradigm. Unlike prior works that typically restrict ensemble depth to shallow meta-levels, RocketStack systematically explores recursive stacking up to level 10. By integrating dynamic model pruning based on OOF scores with selectively applied feature compression strategies, RocketStack sustains performance gains across levels while mitigating the feature inflation and runtime burden traditionally associated with deep ensembling. Evaluation across 33 datasets, spanning both binary and multi-class tasks, revealed consistent accuracy improvements alongside significant runtime savings under periodic feature selection. These findings underscore the feasibility of extending ensemble depth to previously unexplored levels without compromising tractability, setting the stage for future research in high-capacity, interpretable ensembling models.

Beyond raw performance, RocketStack follows a design philosophy that differs from typical AutoML systems. AutoML pipelines emphasise exhaustive hyper-parameter and model search, often at the cost of transparency \cite{freitas2019automl}. In contrast, RocketStack exposes depth, pruning, and compression as controllable levers: each meta-level is built and evaluated explicitly, so researchers can observe how structural choices affect accuracy, runtime, and feature growth. This layer-wise transparency supports interpretability and resource budgeting, capabilities that are increasingly required for explainable and tractable learning systems.


\subsection{Effectiveness of model score randomization prior to pruning across levels}
One of the most compelling findings of this study is the positive effect of injecting mild Gaussian noise into OOF performance scores during model pruning. In RocketStack, this strategy was explored by perturbing OOF-based scores before applying percentile-based pruning at each ensemble level, thereby influencing which learners are retained for the next stage. Notably, this stochastic pruning scheme particularly with light noise ($\lambda = 0.05$) yielded consistent performance gains in configurations with periodic feature selection. This trend was most evident in binary classification using periodic SFE, where OOF-light randomization surpassed the deterministic OOF-strict baseline beginning at level-7 (\autoref{tab:bestvsbest}). Similarly, in the multi-class setting, attention-based periodic compression benefited from the same noise-induced variation. These observations suggest a regularization-like mechanism, where slight uncertainty in pruning prevents early overcommitment to narrowly superior models and fosters ensemble diversity at deeper levels. Conceptually, this is analogous to Dropout in neural networks \cite{srivastava2014dropout}: Dropout multiplies activations by a Bernoulli mask, injecting noise that discourages co-adaptation of neurons; in the present architecture, mild Gaussian noise is added to OOF scores, thereby discouraging premature commitment to a single model subset. Both forms of stochastic perturbation act as regularisers and promote robustness. The resulting long-term performance stability supports the view that controlled noise can improve generalization in deep ensembling contexts.

\subsection{Impact of feature compression frequency on recursive stacking}
Feature compression applied at every level leads to irregular performance fluctuations without exhibiting consistent improvement across stacking depth, in contrast to periodic selection which shows gradual and stable gains. This difference likely stems from premature compression disrupting the natural accumulation of intermediate features that would otherwise yield more meaningful reductions. Periodic selection allows for richer feature formation before compression, enhancing the quality of reduced representations. The effect is especially evident in multiclass settings, most notably with SFE, where performance steadily deteriorates under each-level compression, as seen in \autoref{fig:bestvsbest_benchmark}B, while periodic SFE shows clear improvement over levels. Supporting this contrast, statistical trend analyses in \autoref{fig:trendanalysisresults_on_ensembling_depth}A reveal that none of the each-level compression variants display significant positive trends in either binary or multi-class settings, unlike periodic strategies that consistently lead to statistically validated accuracy improvements. These findings suggest that an accumulate-and-release strategy may offer advantages for recursive stacking, echoing observations from the previous work in deep network compression, where premature reductions were found to impair representational capacity and downstream performance \cite{alvarez2017compression}.

\subsection{Disproportionate early performance improvement as a limiting factor in recursive stack ensembling}
A pronounced performance gain is consistently observed between base-level ensembles (level-0) and the first stacked ensemble layer (level-1), described here as the initial performance surge. This substantial improvement remains robust across various configurations tested in this study, including variations in feature compression frequency and model pruning methods, highlighting the utility of integrating base predictions with original features at the initial stacking step. Beyond level-1, subsequent stacking levels in this study produced more modest and fluctuating improvements, indicating diminishing marginal returns. Nonetheless, meaningful performance gains continued through level-10, particularly in configurations retained all meta-features or applied compression periodically, contributing notably to ensemble robustness. This observed asymmetry aligns with the literature's predominant focus on horizontal model diversity rather than deep vertical stacking, emphasizing a possible limitation inherent in recursive stacking architectures.

\subsection{Influence of base-level hyperparameter tuning on deep ensemble stacking}

As demonstrated in Figure \ref{fig:HPO_influence_on_RocketStack}, while applying Bayesian HPO to Level-0 base learners offers an initial performance advantage, this trajectory eventually plateaus and appears to be matched or slightly surpassed by the default configuration by Level-10. Notably, in the binary classification setting, the HPO-applied RocketStack peaks at Level-6 before its long-term growth levels off. This observation suggests that initial sub-optimality, or a degree of predictive fuzziness in the base models, might not necessarily diminish RocketStack's performance over deepening layers. Rather than requiring heavily optimized base learners, the architecture's periodic feature compression and OOF-guided pruning mechanisms may help process and refine these un-tuned signals over time. Because both configurations reach highly competitive Level-10 accuracies, the results present a practical consideration: the substantial computational cost of base-level HPO might not be necessary to achieve strong performance in highly deep stacked ensemble learning. Furthermore, as summarized in Table \ref{tab:RocketStackvsSOTA}, this un-tuned diversity enables RocketStack at level~10 to slightly outperform the SOTA deep tabular models TabNet and Deep Forest, with Bayesian HPO applied, in both binary and multi-class settings. Taken together, these findings support the view that in deep recursive stacking, preserving base-level variance provides a more stable foundation for downstream generalization than optimizing base-level models.

\subsection{Future work}
This study focused on relatively simple tabular datasets to evaluate the generalizability of RocketStack under diverse but manageable exploratory settings. Future work could extend RocketStack beyond simple tabular datasets to more challenging, large-scale benchmarks where deeper stacking is both viable and necessary. Scaling to such settings would benefit from a larger and more heterogeneous model pool, potentially including duplicate architectures with varied hyperparameters to promote structural diversity. As stacking depth increases, further exploration of pruning dynamics becomes essential not only in terms of randomized pruning, but also through temporally informed strategies that aggregate model performance across multiple levels. Such delayed, memory-based pruning (e.g., using moving averages or decay-weighted trends) \cite{exponential_smoothing} could enable new optimization regimes in architectures considerably deeper than ten levels (hyper-deep ensembles), while preserving both performance and efficiency.

\section{Conclusion}

RocketStack presents a scalable, modular architecture for deep recursive stacking that unifies consistent model pruning with level-aware feature fusion and compression. Models are pruned using dynamic percentile thresholds on out-of-fold (OOF) performance, with an optional Gaussian perturbation of OOF scores to reduce premature convergence toward locally dominant learners and to preserve ensemble diversity across depth. Across 33 datasets (23 binary, 10 multi-class), periodic feature selection, including SFE in binary and attention-based selection in multi-class settings, yields a favorable accuracy–runtime trade-off relative to uncompressed stacking, while per-level compression can be overly aggressive. At the deepest levels, RocketStack slightly exceeds established deep tabular baselines. Overall, RocketStack balances increasing predictive performance with adaptive control of feature dimensionality and computational cost, providing a practical foundation for scalable decision fusion as model pools and feature spaces evolve.


\section*{Declaration of competing interest}
The author declares that there are no competing financial interests or personal relationships that could have appeared to influence the work reported in this paper.

\bibliographystyle{elsarticle-harv}
\bibliography{myBibliography_v1.2}

\appendix

\setcounter{table}{0}
\renewcommand{\thetable}{A\arabic{table}}

\section{Additional Tables}

\subsection{Default Model Configurations and Hyperparameters (Binary \& Multi-class)}

\begin{table}[H]
\centering
\scriptsize
\caption{Model configurations and default hyperparameters used for binary and multi-class classification}
\begin{tabular}{|p{3.5cm}|p{7.5cm}|c|c|}
\hline
\textbf{Model} & \textbf{Hyperparameters} & \textbf{Binary} & \textbf{Multi-class} \\
\hline
\texttt{AdaBoost} & \texttt{n\_estimators=50, learning\_rate=1.0, algorithm='SAMME'}
 & Included & Included \\
\hline
\texttt{SVC} & \texttt{C=1.0, kernel='rbf', gamma='auto', probability=True, tol=1e-3, max\_iter=-1}
 & Included & Included \\
\hline
\texttt{Bagging} & \texttt{estimator=DecisionTreeClassifier(), n\_estimators=10, max\_samples=1.0, max\_features=1.0, bootstrap=True} & Included & Included \\
\hline
\texttt{Random Forest} & \texttt{n\_estimators=100, criterion='gini', max\_features='sqrt', bootstrap=True}
 & Included & Included \\
\hline
\texttt{XGBoost} & \texttt{objective='binary:logistic', booster='gbtree', learning\_rate=0.3, max\_depth=6, n\_estimators=100, gamma=0, min\_child\_weight=1, subsample=1, colsample\_bytree=1, reg\_alpha=0, reg\_lambda=1, scale\_pos\_weight=1} & Included & Included \\
\hline
\texttt{LightGBM} & \texttt{boosting\_type='gbdt', num\_leaves=31, max\_depth=-1, learning\_rate=0.1, n\_estimators=100, subsample=1.0, colsample\_bytree=1.0, reg\_alpha=0.0, reg\_lambda=0.0} & Included & Included \\
\hline
\texttt{KNN} & \texttt{n\_neighbors=5, weights='uniform', algorithm='auto', leaf\_size=30, p=2, metric='minkowski'} & Included & Included \\
\hline
\texttt{LDA} & \texttt{solver='lsqr', shrinkage='auto'} & Included & Not-included \\
\hline
\texttt{Extra Trees} & \texttt{n\_estimators=100, criterion='gini', max\_features='sqrt', bootstrap=False}
 & Included & Included \\
\hline
\texttt{Gradient Boosting} & \texttt{loss='log\_loss', learning\_rate=0.1, n\_estimators=100, subsample=1.0, max\_depth=3, criterion='friedman\_mse'}
 & Included & Included \\
\hline
\texttt{RidgeClassifier} & \texttt{alpha=1.0, fit\_intercept=True, max\_iter=None, tol=1e-4, solver='auto'}
 & Included & Included \\
\hline
\texttt{Calibrated Ridge} & \texttt{cv=5, method='sigmoid'} & Included & Not-included \\
\hline
\texttt{Logistic Regression} & \texttt{penalty='l2', C=1.0, solver='lbfgs', max\_iter=1000, fit\_intercept=True, tol=1e-4}
 & Included & Included \\
\hline
\texttt{Calibrated Passive Aggressive} & \texttt{cv=5, method='sigmoid'} & Included & Not-included \\
\hline
\texttt{BernoulliNB} & \texttt{alpha=1.0, force\_alpha=True, binarize=0.0, fit\_prior=True, class\_prior=None} & Included & Not-included \\
\hline
\texttt{GaussianNB} & \texttt{var\_smoothing=1e-9} & Included & Included \\
\hline
\texttt{MLP} & \texttt{hidden\_layer\_sizes=(100, 50), activation='relu', solver='adam', alpha=0.0001, learning\_rate\_init=0.001, max\_iter=1000, verbose=False} & Included & Included \\
\hline
\texttt{HistGradientBoosting} & \texttt{learning\_rate=0.1, max\_iter=100, max\_depth=None, loss='log\_loss'} & Included & Included \\
\hline
\texttt{SGD} & \texttt{loss='log\_loss', penalty='l2', alpha=0.0001, l1\_ratio=0.15, learning\_rate='optimal', eta0=0.0, max\_iter=1000, tol=1e-3, early\_stopping=False} & Included & Not-included \\
\hline
\texttt{Catboost} & \texttt{iterations=1000, learning\_rate=0.03, depth=6, loss\_function='Logloss', silent=True} & Included & Not-included \\
\hline
\end{tabular}
\label{tab:model_hyperparams_framed}
\end{table}

\subsection{Binary classification trend statistics}

\begin{table}[H]
\fontsize{4}{5}\selectfont
\centering
\caption{Summary statistics from the trend analysis of RocketStack ensembling depth across five feature compression strategies in binary classification. Accuracy values (\%) represent means and standard deviations computed across all models, datasets, and 5-fold cross-validation splits.}
\label{tab:averageperformances_of_binaryclassification_merged}

\scalebox{0.72}[1]{
\begin{tabular}{llc@{\hskip 4pt}*{10}{@{\hskip 4pt}c@{\hskip 4pt}}c}
\toprule
\textbf{Feature selection} & \textbf{Individual} & \textbf{L1} & \textbf{L2} & \textbf{L3} & \textbf{L4} & \textbf{L5} & \textbf{L6} & \textbf{L7} & \textbf{L8} & \textbf{L9} & \textbf{L10} & \textbf{Stack of stack} \\
\midrule
Without feature compression & 85.59 ± 1.55 & 87.89 ± 1.54 & 88.09 ± 1.37 & 88.05 ± 1.56 & 88.08 ± 1.47 & 88.19 ± 1.45 & 88.79 ± 1.43 & 88.96 ± 1.44 & 89.20 ± 1.45 & 89.21 ± 1.12 & 89.74 ± 1.05 & 87.17 ± 1.76 \\
SFE each level              & 85.31 ± 2.04 & 87.09 ± 2.13 & 87.96 ± 3.21 & 87.90 ± 2.59 & 87.69 ± 3.14 & 87.40 ± 3.26 & 87.40 ± 3.14 & 86.69 ± 3.65 & 86.06 ± 4.36 & 85.87 ± 3.90 & 84.52 ± 4.86 & 87.28 ± 2.05 \\
SFE periodic (OOF-strict)   & 85.78 ± 1.66 & 87.71 ± 1.58 & 87.73 ± 1.84 & 87.81 ± 1.71 & 88.56 ± 2.12 & 88.49 ± 2.15 & 88.46 ± 2.05 & 88.36 ± 2.43 & 88.35 ± 1.89 & 87.84 ± 1.87 & 88.22 ± 1.74 & 87.18 ± 1.98 \\
SFE periodic (OOF-light rand.) & 85.04 ± 2.01 & 86.94 ± 1.96 & 87.41 ± 1.74 & 87.49 ± 1.64 & 88.86 ± 1.53 & 88.61 ± 1.44 & 88.93 ± 1.47 & 88.72 ± 1.64 & 88.88 ± 1.61 & 88.67 ± 1.57 & 88.91 ± 1.44 & 86.79 ± 2.08 \\
SFE periodic (OOF-moderate rand.) & 85.16 ± 2.02 & 86.82 ± 2.10 & 86.55 ± 1.88 & 86.99 ± 1.55 & 88.71 ± 1.57 & 88.73 ± 1.50 & 88.56 ± 1.53 & 88.74 ± 1.33 & 89.25 ± 1.28 & 89.67 ± 1.22 & 89.61 ± 1.27 & 86.70 ± 2.10 \\
\bottomrule
\end{tabular}
}
\end{table}

\subsection{Multi-class classification trend statistics}

\begin{table}[H]
\fontsize{5}{6}\selectfont
\centering
\caption{Summary statistics from the trend analysis of RocketStack ensembling depth across ten feature compression strategies in multi-class classification. Accuracy values (\%) represent means and standard deviations computed over all models, datasets, and 5-fold cross-validation splits.}
\label{tab:averageperformances_of_multiclassclassification_merged}

\scalebox{0.7}[1]{
\begin{tabular}{l@{\hskip 8pt}l@{\hskip 8pt}c@{\hskip 8pt}c@{\hskip 8pt}c@{\hskip 8pt}c@{\hskip 8pt}c@{\hskip 8pt}c@{\hskip 8pt}c@{\hskip 8pt}c@{\hskip 8pt}c@{\hskip 8pt}c@{\hskip 8pt}c@{\hskip 8pt}}
\toprule
\textbf{Feature selection} & \textbf{Individual} & \textbf{L1} & \textbf{L2} & \textbf{L3} & \textbf{L4} & \textbf{L5} & \textbf{L6} & \textbf{L7} & \textbf{L8} & \textbf{L9} & \textbf{L10} & \textbf{Stack of stack} \\
\midrule
Without feature compression & 85.17 ± 1.51 & 91.83 ± 1.47 & 92.93 ± 1.23 & 93.98 ± 1.16 & 94.20 ± 0.82 & 94.63 ± 0.97 & 94.83 ± 0.92 & 94.79 ± 0.82 & 94.57 ± 0.74 & 94.39 ± 0.66 & 94.51 ± 0.61 & 92.06 ± 1.27 \\
SFE each level              & 84.62 ± 1.37 & 92.40 ± 1.04 & 92.70 ± 1.38 & 91.68 ± 2.86 & 90.78 ± 3.15 & 88.46 ± 5.13 & 87.26 ± 4.96 & 84.02 ± 5.96 & 81.17 ± 5.38 & 77.00 ± 6.98 & 75.20 ± 6.52 & 92.52 ± 1.09 \\
SFE periodic (OOF-strict)   & 85.84 ± 1.51 & 93.23 ± 1.10 & 93.73 ± 1.15 & 94.17 ± 1.17 & 94.23 ± 1.37 & 94.39 ± 1.03 & 94.54 ± 0.97 & 94.18 ± 1.27 & 94.28 ± 1.03 & 94.19 ± 1.27 & 94.43 ± 1.18 & 93.71 ± 1.12 \\
Autoencoders each level (2L) & 83.21 ± 1.43 & 92.63 ± 1.32 & 87.79 ± 4.94 & 88.14 ± 4.75 & 88.67 ± 4.23 & 88.65 ± 4.13 & 88.10 ± 3.81 & 87.60 ± 3.58 & 87.07 ± 3.70 & 87.35 ± 3.37 & 87.01 ± 1.66 & 89.87 ± 3.86 \\
Autoencoders periodic (2L)  & 83.76 ± 1.46 & 93.03 ± 0.93 & 93.46 ± 1.14 & 93.64 ± 1.07 & 88.36 ± 4.53 & 88.36 ± 4.41 & 88.71 ± 4.08 & 89.24 ± 4.21 & 90.05 ± 3.82 & 90.47 ± 3.40 & 89.67 ± 3.97 & 91.33 ± 2.78 \\
Autoencoders periodic (3L)        & 88.47 ± 1.45 & 93.15 ± 1.12 & 93.79 ± 1.18 & 94.16 ± 1.19 & 88.59 ± 5.24 & 88.49 ± 4.90 & 89.76 ± 4.58 & 89.09 ± 4.83 & 89.96 ± 4.52 & 90.61 ± 3.83 & 91.30 ± 3.60 & 91.42 ± 3.59 \\
Attention each level              & 85.91 ± 1.45 & 93.63 ± 1.08 & 94.00 ± 1.43 & 94.09 ± 1.41 & 94.08 ± 1.20 & 94.28 ± 1.14 & 94.19 ± 1.19 & 94.25 ± 1.30 & 94.33 ± 1.15 & 93.49 ± 1.24 & 93.57 ± 0.94 & 93.88 ± 1.02 \\
Attention periodic (OOF-strict)   & 85.64 ± 1.52 & 92.88 ± 1.15 & 93.26 ± 1.08 & 93.57 ± 1.17 & 93.98 ± 0.96 & 94.25 ± 0.91 & 94.19 ± 0.94 & 94.40 ± 0.81 & 94.16 ± 0.79 & 94.61 ± 0.78 & 94.75 ± 0.65 & 93.23 ± 1.06 \\
Attention periodic (OOF-light)    & 85.39 ± 1.40 & 93.18 ± 1.20 & 93.64 ± 1.36 & 94.29 ± 1.20 & 94.23 ± 1.27 & 94.29 ± 1.17 & 94.58 ± 0.74 & 94.23 ± 0.96 & 94.63 ± 0.92 & 93.97 ± 0.76 & 94.83 ± 0.50 & 93.68 ± 1.10 \\
Attention periodic (OOF-moderate) & 85.31 ± 1.49 & 93.00 ± 1.15 & 93.47 ± 1.07 & 93.82 ± 1.03 & 93.67 ± 1.10 & 93.50 ± 0.92 & 93.72 ± 0.87 & 93.15 ± 1.04 & 93.53 ± 0.98 & 93.88 ± 0.52 & 93.78 ± 0.53 & 93.51 ± 1.12 \\
\bottomrule
\end{tabular}
}
\end{table}


\subsection{Accuracy Trend Statistics (Binary \& Multi-class)}

\begin{table}[H]
\fontsize{7}{8}\selectfont
\centering
\caption{Statistical analysis of accuracy trends across RocketStack ensembling levels in binary and multi-class classification. Reported p-values are derived from linear mixed models (LMMs) without quadratic terms. Bonferroni correction is applied across settings. The label \textit{periodic} refers to feature selection applied specifically at levels 3, 6, and 9.}
\label{tab:trend_analysis}

\begin{tabularx}{\textwidth}{llXX}
\toprule
\multirow{2}{*}{\textbf{Classification type}} 
& \multirow{2}{*}{\textbf{Feature selection}} 
& \multicolumn{2}{c}{\textbf{\emph{p}-value}} \\
\cmidrule(lr){3-4}
& & Uncorr. & Bonferroni-corr. \\
\midrule
Binary & Without feature compression & $<.001$ & $<.001$ \\
Binary & SFE each level & .027 & .411 \\
Binary & SFE periodic(OOF-strict) & $<.001$ & .004 \\
Binary & SFE periodic (OOF-light randomized) & $<.001$ & $<.001$ \\
Binary & SFE periodic (OOF-moderate randomized) & $<.001$ & $<.001$ \\
\midrule
Multi-class & Without feature compression & $<.001$ & $<.001$ \\
Multi-class & SFE each level & $<.001$ & $<.001$ \\
Multi-class & SFE periodic (OOF-strict) & .001 & .013 \\
Multi-class & Autoencoders each level, 2L & .587 & 1.000 \\
Multi-class & Autoencoders periodic, 2L & .437 & 1.000 \\
Multi-class & Autoencoders periodic, 3L & .464 & 1.000 \\
Multi-class & Attention each level & .006 & .087 \\
Multi-class & Attention periodic (OOF-strict) & $<.001$ & .002 \\
Multi-class & Attention periodic (OOF-light randomized) & $<.001$ & .003 \\
Multi-class & Attention periodic (OOF-moderate randomized) & .002 & .026 \\
\bottomrule
\end{tabularx}
\end{table}

\subsection{Benchmark comparison of grand-averaged best performing model accuracy}

\begin{table}[H]
\fontsize{6}{8}\selectfont
\centering
\caption{Benchmark comparison of the grand averaged best performing model performances (Accuracy [\%]) over datasets per ensembling level.} 
\label{tab:bestvsbest}

\begin{tabular}{llc@{\hskip 4pt}*{10}{@{\hskip 4pt}c@{\hskip 4pt}}}
\toprule
\textbf{Classification} & \textbf{Feature selection} & \shortstack{\textbf{Baseline} \\ \textbf{ensemble}} & \textbf{L1} & \textbf{L2} & \textbf{L3} & \textbf{L4} & \textbf{L5} & \textbf{L6} & \textbf{L7} & \textbf{L8} & \textbf{L9} & \textbf{L10} \\

\midrule
Binary & Without feature compression & 88.46 & 90.35 & 90.35 & 90.22 & 90.09 & 90.40 & 90.19 & 91.56 & 93.68 & 93.17 & 97.69\\
Binary & Each level SFE & 88.46 & 90.36 & 90.56 & 89.92 & 91.00 & 91.73 & 93.76 & 93.06 & 95.07 & 94.41 & 90.41\\
Binary & Periodic SFE (No rand.) & 88.46 & 90.13 & 90.12 & 90.05 & 90.91 & 89.37 & 92.33 & 93.22 & 93.08 & 91.69 & 91.94\\
Binary & Periodic SFE (Rand-$\lambda=0.1$) & 88.46 & 90.23 & 90.2 & 90.06 & 90.33 & 89.85 & 89.74 & 97.08 & 97.08 & 97.08 & 97.08\\
Binary & Periodic SFE (Rand-$\lambda=0.05$) & 88.46 & 90.28 & 90.4 & 90.25 & 90.08 & 90.42 & 90.8 & 90.24 & 91.0 & 91.22 & 93.54\\
\midrule
Multi-class & Without feature compression & 92.49 & 94.76 & 95.74 & 95.62 & 95.9 & 95.99 & 96.26 & 96.3 & 96.4 & 96.24 & 96.36\\
Multi-class & Each level SFE & 92.49 & 93.62 & 94.39 & 93.92 & 92.99 & 91.6 & 90.22 & 87.4 & 84.58 & 83.62 & 80.84\\
Multi-class & Periodic SFE (No rand.) & 92.49 & 94.42 & 95.44 & 95.5 & 95.29 & 95.46 & 95.7 & 97.68 & 97.88 & 98.04 & 97.76\\
Multi-class & Each level Autoenc. 2L & 92.49 & 94.26 & 91.33 & 91.34 & 94.19 & 93.4 & 93.5 & 92.85 & 92.33 & 92.83 & 97.86\\
Multi-class & Periodic Autoenc. 2L & 92.49 & 94.42 & 95.41 & 95.19 & 94.47 & 92.27 & 92.27 & 94.56 & 97.87 & 97.87 & 94.78\\
Multi-class & Periodic Autoenc. 3L & 92.49 & 94.5 & 95.42 & 95.36 & 95.04 & 95.03 & 95.03 & 95.36 & 98.08 & 98.51 & 98.16\\
Multi-class & Each level Attention & 92.49 & 94.88 & 95.25 & 95.72 & 95.36 & 97.68 & 97.44 & 97.83 & 97.96 & 97.85 & 97.26\\
Multi-class & Periodic Attention (No rand.) & 92.49 & 94.22 & 95.23 & 95.15 & 95.16 & 96.2 & 96.05 & 97.62 & 97.68 & 98.51 & 98.19\\
Multi-class & Periodic Attention (Rand-$\lambda=0.1$) & 92.49 & 94.59 & 95.69 & 98.12 & 97.75 & 97.84 & 97.98 & 97.64 & 97.72 & 98.01 & 98.14\\
Multi-class & Periodic Attention (Rand-$\lambda=0.05$) & 92.49 & 94.35 & 95.31 & 95.48 & 95.58 & 95.77 & 98.36 & 98.16 & 98.24 & 98.12 & 98.6\\
\bottomrule

\end{tabular}
\end{table}



\subsection{(a) Normalized Runtime and (b) Mean Feature Count Across Levels (L1–L10)}

\begin{table}[H]
\fontsize{6.5}{8}\selectfont
\centering
\caption{Comparison of normalized runtime (a) and averaged number of features (b) across different feature selection and model pruning strategies in binary and multi-class classification. Runtime values are normalized to the [0–1] scale across all configurations. Metrics are reported per ensemble level from L1 to L10.}
\label{tab:runtime_and_numberoffeatures}

\begin{subtable}[t]{\textwidth}
\centering
\caption{Runtime comparison across feature selection \& model pruning strategies on binary and multi-class settings.}
\label{tab:runtimecomparisons}

\begin{tabular}{llc@{\hskip 4pt}*{10}{@{\hskip 4pt}c@{\hskip 4pt}}}
\toprule
\textbf{Classification} & \textbf{Feature selection} & \textbf{Individual} & \textbf{L1} & \textbf{L2} & \textbf{L3} & \textbf{L4} & \textbf{L5} & \textbf{L6} & \textbf{L7} & \textbf{L8} & \textbf{L9} & \textbf{L10} \\
\midrule
Binary & Without feature compression & 0.031 & 0.167 & 0.179 & 0.192 & 0.208 & 0.226 & 0.245 & 0.267 & 0.291 & 0.316 & 0.344\\
Binary & Each level SFE & 0.031 & 0.164 & 0.196 & 0.222 & 0.248 & 0.273 & 0.298 & 0.321 & 0.344 & 0.367 & 0.389\\
Binary & Periodic SFE & 0.031 & 0.167 & 0.179 & 0.193 & 0.23 & 0.236 & 0.243 & 0.272 & 0.276 & 0.281 & 0.308\\
\midrule
Multi-class & Without feature compression & 0.023 & 0.124 & 0.162 & 0.223 & 0.306 & 0.411 & 0.52 & 0.651 & 0.776 & 0.885 & 1.00\\
Multi-class & Each level SFE & 0.021 & 0.091 & 0.144 & 0.185 & 0.237 & 0.291 & 0.346 & 0.403 & 0.466 & 0.531 & 0.593\\
Multi-class & Periodic SFE & 0.022 & 0.098 & 0.153 & 0.242 & 0.372 & 0.455 & 0.518 & 0.581 & 0.636 & 0.697 & 0.754\\
Multi-class & Each level Autoenc. 2L & 0.019 & 0.082 & 0.121 & 0.147 & 0.17 & 0.186 & 0.2 & 0.216 & 0.226 & 0.232 & 0.237\\
Multi-class & Periodic Autoenc. 2L & 0.019 & 0.084 & 0.118 & 0.184 & 0.269 & 0.302 & 0.35 & 0.407 & 0.431 & 0.465 & 0.507\\
Multi-class & Periodic Autoenc. 3L & 0.019 & 0.086 & 0.122 & 0.183 & 0.265 & 0.304 & 0.348 & 0.403 & 0.422 & 0.45 & 0.485\\
Multi-class & Each level Attention & 0.022 & 0.098 & 0.247 & 0.317 & 0.368 & 0.414 & 0.456 & 0.49 & 0.521 & 0.532 & 0.543\\
Multi-class & Periodic Attention & 0.02 & 0.088 & 0.119 & 0.176 & 0.257 & 0.283 & 0.315 & 0.362 & 0.376 & 0.403 & 0.439\\
\bottomrule

\end{tabular}
\end{subtable}

\vskip 3mm

\begin{subtable}[t]{\textwidth}
\centering
\caption{Averaged number of features on ensembling depth on binary and multi-class settings.}
\label{tab:number_of_features}

\begin{tabular}{llc@{\hskip 4pt}*{10}{@{\hskip 4pt}c@{\hskip 4pt}}}
\toprule
\textbf{Classification} & \textbf{Feature selection} & \textbf{L1} & \textbf{L2} & \textbf{L3} & \textbf{L4} & \textbf{L5} & \textbf{L6} & \textbf{L7} & \textbf{L8} & \textbf{L9} & \textbf{L10} \\
\midrule
Binary & Without feature compression & 51.57 & 69.77 & 86.66 & 102.52 & 117.40 & 131.31 & 144.26 & 156.24 & 167.26 & 177.3\\
Binary & Each level SFE & 51.57 & 5.7 & 4.23 & 3.77 & 3.73 & 3.63 & 3.18 & 3.43 & 3.15 & 3.05\\
Binary & Periodic SFE & 51.57 & 69.77 & 86.66 & 7.55 & 22.41 & 36.28 & 5.83 & 17.75 & 28.70 & 5.97\\
\midrule
Multi-class & Without feature compression & 145.56 & 239.44 & 325.64 & 404.61 & 477.16 & 543.61 & 604.76 & 661.67 & 714.28 & 762.91\\
Multi-class & Each level SFE & 145.56 & 43.94 & 22.3 & 19.88 & 17.84 & 15.46 & 14.5 & 12.34 & 10.5 & 3.12\\
Multi-class & Periodic SFE & 145.56 & 242.3 & 331.9 & 75.26 & 151.96 & 221.54 & 43.48 & 101.1 & 141.8 & 31.5\\
Multi-class & Each level Autoenc. 2L & 145.56 & 80.4 & 56.2 & 45.6 & 39.88 & 35.56 & 32.08 & 28.76 & 23 & 13.26\\
Multi-class & Periodic Autoenc. 2L & 145.56 & 242.5 & 332.02 & 137.82 & 213.64 & 281.96 & 114.32 & 170.7 & 210.64 & 52.8\\
Multi-class & Periodic Autoenc. 3L & 145.56 & 242.5 & 331.7 & 137.46 & 212.9 & 281.08 & 113.92 & 170.14 & 220 & 63.2\\
Multi-class & Each level Attention & 145.56 & 59.92 & 37.14 & 29.68 & 26.36 & 23.7 & 21.58 & 19.36 & 15.8 & 8.98\\
Multi-class & Periodic Attention & 145.56 & 242.1 & 331.24 & 103.42 & 178.48 & 246.1 & 76.84 & 133 & 166.6 & 37.86\\
\bottomrule

\end{tabular}
\end{subtable}

\end{table}


\clearpage
\renewcommand{\thefigure}{S\arabic{figure}} 
\setcounter{figure}{0}   

\begin{center}
\textbf{\large Supplementary Material}
\end{center}

\begin{figure}[H]
 \centering 
 \includegraphics[width=0.8\textwidth,keepaspectratio=false]{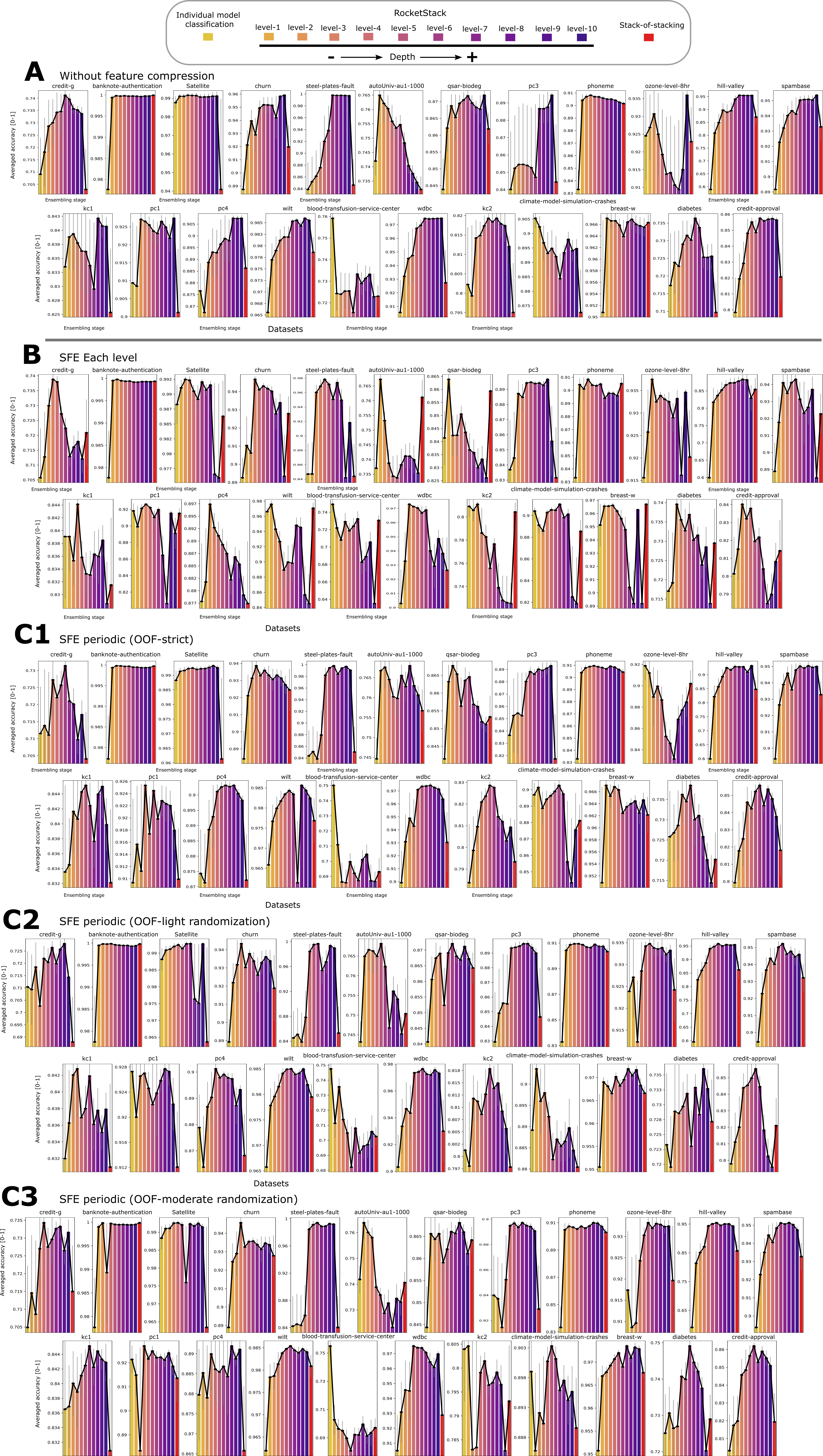}
 \caption{Accuracy (\%) across ensembling depths for each individual binary classification dataset for each RocketStack variant.} 
 \label{fig:blendensembleresults_binary} 
\end{figure}

\begin{figure*}
 \centering 
 \includegraphics[width=1\textwidth,keepaspectratio=false]{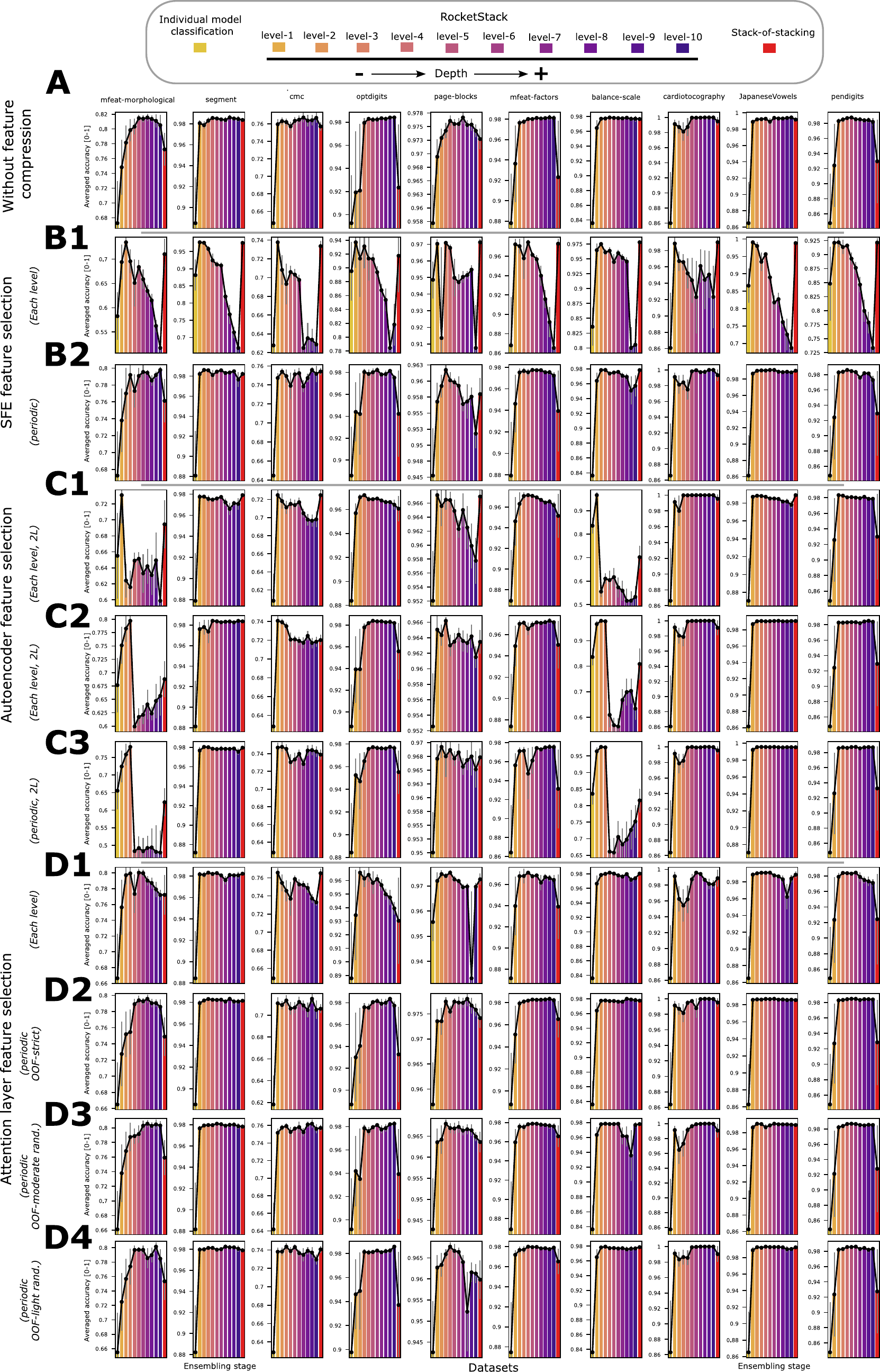}
 \caption{Accuracy (\%) across ensembling depths for each individual multi-class classification dataset for each RocketStack variant. \textit{periodic} refers to level 3, 6 and 9}. 
 \label{fig:blendensembleresults_multiclass} 
\end{figure*}

\end{document}